%% file: main.tex
\documentclass{article}
\PassOptionsToPackage{round}{natbib}
 \usepackage[preprint]{neurips_2026}

\usepackage[utf8]{inputenc} 
\usepackage[T1]{fontenc}    
\usepackage{hyperref}       
\usepackage{url}            
\usepackage{booktabs}       
\usepackage{amsfonts}       
\usepackage{nicefrac}       
\usepackage{microtype}      
\usepackage{xcolor}         

\usepackage{amsmath}
\hypersetup{hidelinks}
\usepackage[nohyperlinks]{acronym}
\usepackage[round]{natbib}

\input{elements/mapes_old}
\input{elements/mapes}

\usepackage{algorithm}
\usepackage[capitalize,noabbrev]{cleveref}
\usepackage{wrapfig}
\usepackage{multicol} 
\usepackage[frozencache,cachedir=minted-cache]{minted}
\usepackage{listings}
\usepackage{xcolor}
\usepackage{xfp}
\usepackage{algpseudocode}
\usepackage{placeins}
\usepackage{caption}
\usepackage{multirow}
\usepackage{subcaption}
\usepackage{booktabs}
\usepackage[disable,textsize=tiny]{todonotes}
\newcommand{\ourwork}{EnergyLens}

\title{EnergyLens: Predictive Energy-Aware Exploration\\for Multi-GPU LLM Inference Optimization}
\author{Zhiye Song \\
  Massachusetts Institute of Technology\\
  Cambridge, MA 02139 \\
  \texttt{zhiye@mit.edu} \\
  \And
  Kyungmi Lee \\
  Massachusetts Institute of Technology\\
  Cambridge, MA 02139 \\
  \texttt{kyungmi@mit.edu} \\
  \And
  Eun Kyung Lee \\
  IBM Research\\
  Yorktown Heights, NY 10598 \\
  \texttt{eunkyung.lee@us.ibm.com} \\
  \And
  Xin Zhang \\
  IBM Research\\
  Yorktown Heights, NY 10598 \\
  \texttt{xzhang@us.ibm.com} \\
  \And
  Tamar Eilam \\
  IBM Research\\
  Yorktown Heights, NY 10598 \\
  \texttt{eilamt@us.ibm.com} \\
  \And
  Anantha P. Chandrakasan \\
  Massachusetts Institute of Technology\\
  Cambridge, MA 02139 \\
}

\begin{document}

\maketitle

\begin{abstract}
\input{abstract}

\end{abstract}

\input{text}

\begin{ack}
This work was supported by the MIT-IBM Watson AI Lab and by the Natural Sciences and Engineering Research Council of Canada (NSERC) Postgraduate Scholarship program.
\end{ack}
\newpage

\nocite{nvidiaNCCLTests}

{
\small

\input{main.bbl}
}


\appendix
\crefalias{section}{appendix}
\input{appendices}

\clearpage
\newpage
\section{Abbreviations}

\input{elements/abr}



\end{document}

%% file: elements/mapes_old.tex
\newcommand{\oldprefillmultiE}{12.32\%}
\newcommand{\oldprefillmultiT}{12.78\%}
\newcommand{\olddecodemultiE}{12.29\%}
\newcommand{\olddecodemultiT}{28.15\%}
\newcommand{\decodemultiTLi}{210.59\%}
\newcommand{\decodemultiTNeusight}{25.69\%}

\usepackage{xfp}

\newcommand{\fuseprefill}{4.37\%}
\newcommand{\fusedecode}{8.18\%}

%% file: elements/mapes.tex
\newcommand{\prefillsingleE}{11.31\%}
\newcommand{\prefillsingleT}{11.30\%}
\newcommand{\decodesingleE}{12.87\%}
\newcommand{\decodesingleT}{15.21\%}

\newcommand{\prefillmultiE}{12.18\%}
\newcommand{\prefillmultiT}{12.39\%}
\newcommand{\decodemultiE}{12.86\%}

\newcommand{\decodemultiT}{25.45\%}

\def\prefillrednum{20.87}
\def\decoderednum{92.63}

\newcommand{\prefillred}{\prefillrednum\%}
\newcommand{\decodered}{\decoderednum\%}

\newcommand{\prefillvar}{$1.47\times$}
\newcommand{\decodevar}{$52.9\times$}

\newcommand{\qwenPrefillE}{12.13\%}
\newcommand{\qwenPrefillT}{12.17\%}

\newcommand{\qwenDecodeE}{13.46\%}
\newcommand{\qwenDecodeT}{16.83\%}

\newcommand{\qwenTEPE}{9.25\%}
\newcommand{\qwenTEPT}{21.85\%}
\newcommand{\qwenTEPdecodeE}{13.19\%}
\newcommand{\qwenTEPdecodeT}{27.16\%}

\newcommand{\overlapE}{12.97\%}
\newcommand{\overlapT}{11.18\%}

\newcommand{\commPct}{23\%}

\newcommand{\FullExactRecall}{69\%}
\newcommand{\OverlapExactRecall}{20\%}

%% file: abstract.tex
We present \ourwork{}, an end-to-end framework for energy-aware \ac{LLM} inference optimization.
As \ac{LLM}s scale, predicting and reducing their energy footprint has become critical for sustainability and datacenter operations, yet existing approaches either require production-level code and expensive profiling or fail to accurately capture multi-GPU energy behavior.
As a result, practitioners lack tools for deciding which optimizations to prioritize and for selecting among existing deployment configurations when exhaustive profiling is impractical.
\ourwork{} addresses this gap with an intuitive einsum-based interface that captures \ac{LLM} specifications including fusion, parallelism, and compute-communication overlap, combined with load-imbalance-aware MoE modeling and an empirically driven communication energy model for multi-GPU settings.
We validate \ourwork{} on Llama3 and Qwen3-MoE across tensor-parallel and expert-parallel configurations, achieving \acp{MAPE} between \qwenTEPE{} and \qwenTEPdecodeE{} for multi-GPU prefill and decode energy, and \overlapE{} across SM allocations for Megatron-style overlap.
Our energy-driven exploration reveals up to \prefillvar{} and \decodevar{} energy variation across configurations in prefill and decode efficiency and motivates distributed serving. We further show that compute-communication overlap is difficult to optimize with intuition alone, but EnergyLens correctly identifies Pareto-optimal overlap configurations.

%% file: text.tex
\section{Introduction}
The arrival of generative AI, with \acp{LLM} at its core, has transformed machine learning. As \acp{LLM} scale to hundreds of billions of parameters and serve an ever-growing user base, their compute demands translate into substantial energy costs and carbon footprints~\citep{faiz_llmcarbon_2023, mlperf_power, henderson2022systematic, nlp_energy}. Understanding and optimizing inference energy has therefore become critical for both sustainability and datacenter provisioning and serving efficiency.

Meeting this challenge requires energy models that are useful throughout the \ac{LLM} inference development cycle, not only after deployment. In practice, practitioners need energy estimates in two settings: before implementation, to decide which optimizations such as fusion, parallelism, or compute-communication overlap to prioritize, and during deployment, to select among configurations as workloads and latency targets vary, when exhaustive profiling is too expensive. A model for space exploration is especially important in settings where the Pareto front is difficult to infer from intuition alone.
Existing approaches each address part of this problem but fall short for energy-aware \ac{LLM} optimization. Direct measurement is accurate, but requires GPU access and significant engineering effort to implement every configuration, making broad exploration slow and costly. Coarse holistic methods, including \ac{TDP}-based estimation~\citep{mlco2, bloom_co2}, are easy to apply but miss the strong power variations between \ac{LLM} inference phases; in our Llama3-70B measurements on A100 GPUs, TDP-based method can overestimate decode energy by up to 60\%. LLMCO2~\citep{llmco2} predicts power but does not natively support fused implementations or \ac{MoE} architectures, nor does it provide the energy attribution needed to understand optimization opportunities. Kernel-level predictors such as EnergAIzer~\citep{energaizer} provide compute-kernel attribution, but do not capture the multi-GPU effects that increasingly dominate practical \ac{LLM} serving, including parallelism and compute-communication overlap.

These limitations are especially important because modern serving stacks increasingly combine distributed execution with architectural diversity. Models are commonly deployed across multiple GPUs to satisfy memory and latency constraints~\citep{shoeybi_megatron-lm_2019, aminabadi_deepspeed-_2022}, while \ac{MoE} models introduce additional expert-parallel execution and routing imbalance that change both kernel shapes and per-GPU work distribution. As multi-GPU parallelism places greater demand on the network, accurate communication energy modeling becomes essential. Otherwise, significant optimization opportunities can be missed. Take Llama3-70B with 8-GPU tensor parallel as an example, communication accounts for \commPct{} of total energy. Compute-communication overlap~\citep{nemo_overlap, hong2025flux} further complicates latency-energy trade-offs in ways that intuition and existing tools cannot capture. The resulting configuration space spans fusion strategies, parallelism options, overlap settings, batch sizes, each interacting with sequence lengths and latency targets. This combinatorial explosion makes exhaustive profiling impractical.

\ourwork{} addresses this gap by enabling rapid exploration of the energy-latency trade-off space without requiring existing code or GPU access. Developers specify \ac{LLM} architectures with fusion, parallelism, and overlap settings in a few lines of code using our simple interface. The framework parses these specifications and produces detailed energy breakdowns and energy-latency trade-offs with empirically driven, overlap-aware communication modeling and load-imbalance-aware \ac{MoE} modeling. For example, comparing two fusion settings for Llama3-70B prefill across 7 batch sizes and 32 input lengths would require profiling on an 8-GPU server for 12 hours, plus substantial engineering effort. Using \ourwork{}, generating the corresponding results and visualizations only takes 15 minutes.

Our main contributions are as follows:
\begin{itemize}
    \item We introduce \ourwork{}, a framework that lets practitioners describe dense and \ac{MoE} \ac{LLM} inference in a few lines of einsum-based interface with parallelism and overlap annotations, and predicts detailed energy and latency breakdowns without requiring execution traces, production implementations, or GPU access.
    
    \item We develop a distributed energy modeling stack that combines empirically driven communication energy modeling, overlap-aware aggregation, and load-imbalance-aware \ac{MoE} modeling, enabling support for tensor parallelism, expert parallelism, and Megatron-style compute-communication overlap.

    \item We validate \ourwork{} on Llama3 and Qwen3-MoE, achieving \acfp{MAPE} between \qwenTEPE{} and \qwenTEPdecodeE{} for multi-GPU energy and \overlapE{} for overlap evaluations. We show that these errors are small enough to recover the best deployment configurations, including non-intuitive overlap-related trade-offs.

\end{itemize}

\section{Related Work}

\subsection{Kernel-Level Performance Modeling}
Several kernel performance models \citep{neusight, li2023, energaizer} offer faster alternatives to cycle-accurate GPU simulation \citep{kandiah_accelwattch_2021}. We adopt EnergAIzer \citep{energaizer} as our primary compute kernel backend because it provides both latency and power estimates; our framework's modular design also supports alternative latency backends including NeuSight \citep{neusight} and Li et al. \citep{li2023}.

\subsection{LLM Performance Modeling}
Existing \ac{LLM} performance modeling spans both training \citep{lumos, TrioSim, astrasim2} and inference \citep{neusight, llmservingsim, vidur}, but these tools share two key limitations: they focus on latency and throughput while neglecting energy, and most require execution traces from existing implementations, precluding early-stage design exploration. LLMCO2 \citep{llmco2} prioritizes energy, but does not produce the energy breakdowns needed to facilitate optimization opportunities. Moreover, the graph neural network assumes the unfused version of the standard transformer (e.g. Llama2). Applying operation fusion alters this topology, requiring profiling of the fused implementation and retraining the model, neither of which is addressed in the paper.
LLMServingSim v2.0 \citep{llmservingsim2} adds a coarse three-state power model to its performance-focused model, but uses a single power number for active GPUs, missing the diverse power profiles such as between prefill and decode phases. \ourwork{} addresses all these gaps: energy-centric modeling, breakdown visibility, and execution-trace-free input.

\subsection{Compute-Communication Overlap}
Recent work overlaps tensor-parallel collectives with GEMMs to hide communication latency in distributed training; this is also increasingly explored in the prefill phase of distributed inference~\citep{nemo_overlap, lee2025overlap, hong2025flux}.
Predictive modeling of its energy impact, however, remains unexplored \citep{neusight,llmco2,llmservingsim2}. \citet{lee2025overlap} characterizes overlap empirically but does not predict performance. \citet{hong2025flux} optimizes overlap strategies using a latency model that assumes a fixed SM allocation for communication and linear scaling of GEMM partitions. This latency model deteriorates with varying SM allocations or under smaller workloads.

\section{\ourwork{} Framework}
\label{sec:framework}

\begin{figure}
    \centering
    \includegraphics[width=0.9\linewidth]{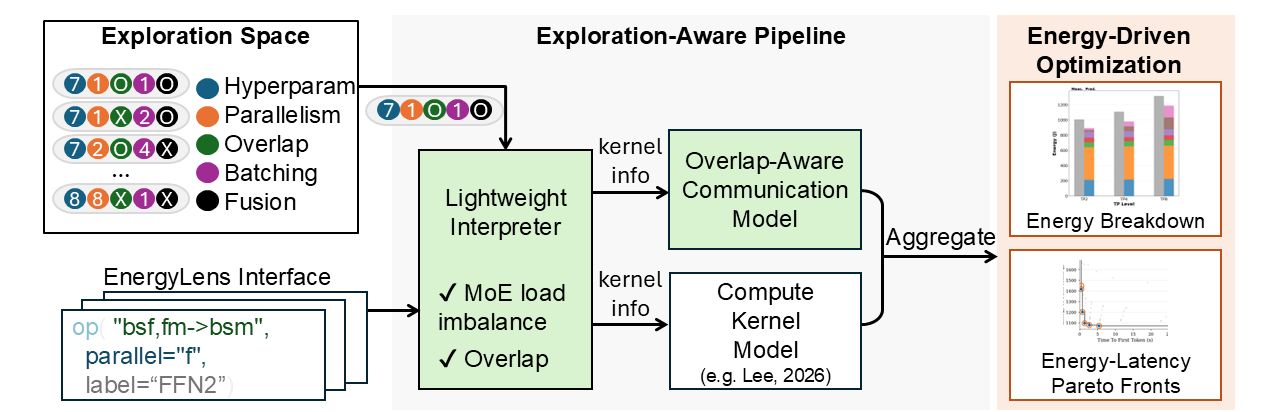}
    \caption{\ourwork{} framework enables energy-latency optimization in the high-dimensional exploration space at early stages of model implementation and deployment.}
    \label{fig:framework}
\end{figure}

\Cref{fig:framework} illustrates the EnergyLens framework. The \emph{EnergyLens Interface} (\Cref{sec:interface}) provides an einsum-based specification for describing LLM architectures, fusion strategies, and parallelism settings. A \emph{Lightweight Interpreter} (\Cref{sec:interpreter}) parses these specifications to extract GEMM dimensions, arithmetic intensity, and required communication kernels. It also supports load imbalance modeling in \ac{MoE} models. The resulting kernel information feeds into an \emph{Overlap-Aware Communication Model} (\Cref{sec:communication}), which captures multi-GPU communication energy, and a \emph{Compute Kernel Model} that leverages EnergAIzer~\citep{energaizer} for single-GPU kernel latency and power estimation, with support for alternative compute latency backends such as NeuSight and Li's model~\citep{neusight,li2023}. The kernel-level results go through an aggregation stage that accounts for compute-communication overlap (\Cref{sec:overlap}) to produce end-to-end energy breakdowns and energy-latency Pareto fronts.

\subsection{\ourwork{} Interface}
\label{sec:interface}
\ourwork{} exposes an einsum-based interface for specifying LLM inference operators, with added lightweight annotations for parallelism and overlap. We choose einsum notation because it is widely used in frameworks such as PyTorch for expressing tensor operations \citep{pytorch_foundation_torcheinsum_nodate}, enabling intuitive descriptions that are easily extended to distributed implementations.

\Cref{lst:qwen} shows a complete fused \ac{MoE} specification in a few lines of code, with symbols defined in \Cref{tab:notation}. Fused and unfused dense model examples can be found in \Cref{sec:app_spec}. For example, the fused QKV projection is written as \texttt{bsm,miKh->bsiKh}, and tensor parallelism is specified by annotating the sharded dimension (e.g., \texttt{parallel="K"}). The same specification covers both prefill and decode: $s$ denotes sequence length in prefill and $s{=}1$ in decode. Model dimensions are supplied at runtime, so one specification can represent models at different scales.

\input{elements/llm_spec_qwen}

Alternative fusion strategies are expressed by modifying only the symbolic operator descriptions. For instance, \Cref{lst:qwen} fuses Q, K, and V projections into one operator (line~\ref{line:qkv}) and combines the gate and up projections (line~\ref{line:gateup}). \ourwork{} also supports distributed execution by annotating the partitioned dimension, e.g., \texttt{parallel="m"} for tensor parallelism and \texttt{CP\_dim="s"} for context parallelism. This compact interface enables rapid exploration of the energy-latency optimization space across \ac{ISL}, \ac{OSL}, batch size, model size, and parallelism.

\subsection{Lightweight Interpreter and Aggregator}
\label{sec:interpreter}

The lightweight interpreter maps each einsum specification to kernel-level features used for energy and latency prediction. For compute operators, it extracts the corresponding \ac{GEMM} dimensions in the standard form $A[M,K]\times B[K,N]=C[M,N]$.
The interpreter also computes arithmetic intensity
to characterize whether an operator is compute- or memory-bound. 

For distributed configurations, the interpreter derives the required communication kernels from the einsum equation and the annotated sharded dimension. For example, if the parallelized dimension is reduced, the operator requires a collective communication step, whose message size is determined from the output tensor shape and data type (details in \Cref{sec:get_comm}).

\begin{figure}[tbhp]
    \centering
    \begin{subfigure}[b]{0.48\linewidth}
        \centering
        \includegraphics[width=0.8\linewidth]{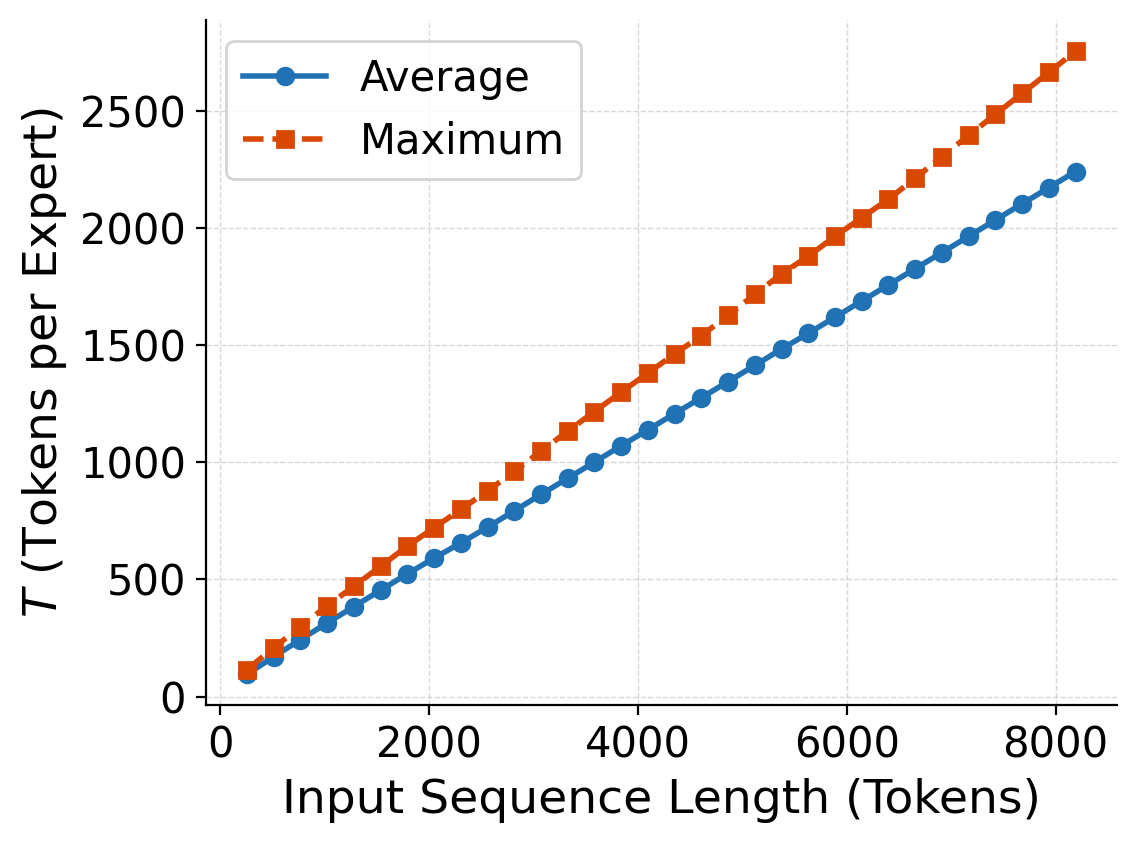}
        \caption{Tokens per expert, $T$, in prefill (batch size = 4, \ac{EP} = 4).}
        \label{fig:prefill_T}
    \end{subfigure}
    \hfill
    \begin{subfigure}[b]{0.48\linewidth}
        \centering
        \includegraphics[width=0.8\linewidth]{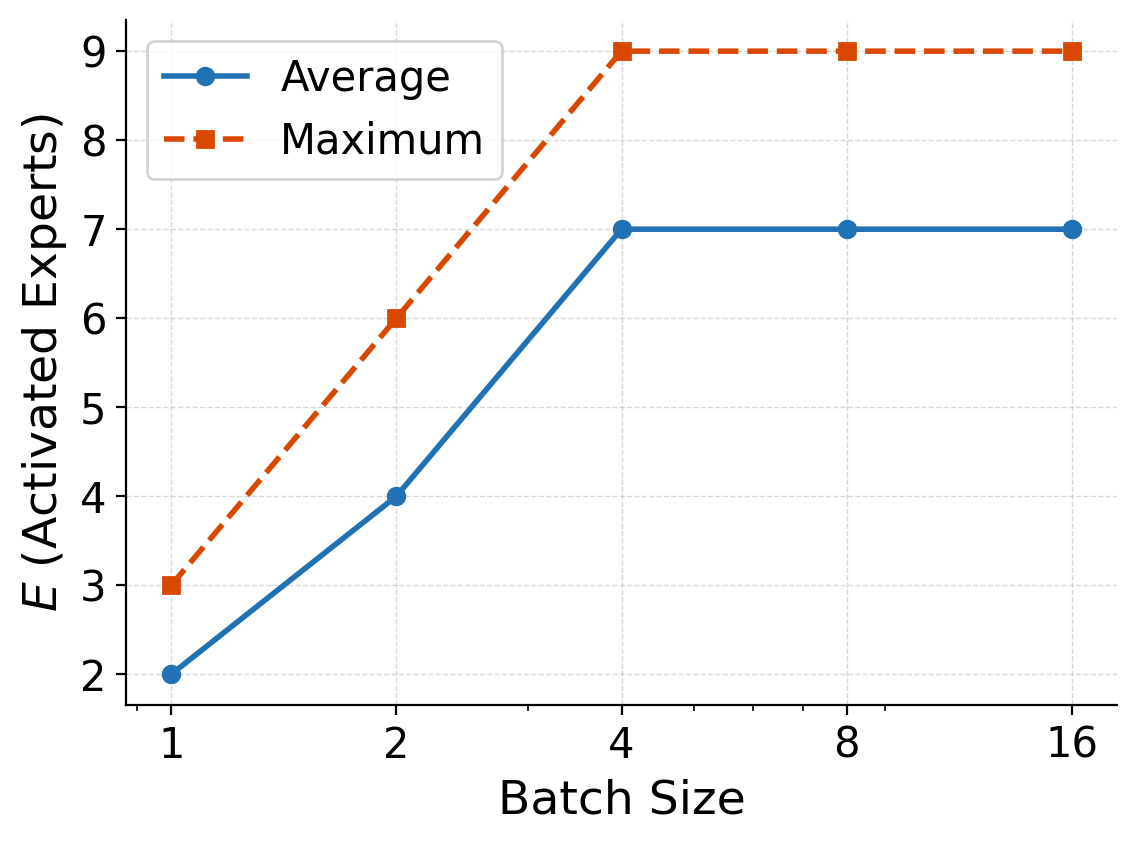}
        \caption{The number of activated experts per GPU, $E$, in decode (\ac{EP} = 4).}
        \label{fig:decode_A}
    \end{subfigure}
    \caption{MoE routing statistics showing the average load and the load of the bottleneck GPU.}
    \label{fig:moe_routing}
\end{figure}

For MoE models, the dimension derivation is more involved, especially when expert-parallel execution depends on how tokens are routed among experts. We therefore introduce two effective quantities: $E$, the number of activated experts per GPU and $T$, the effective number of tokens per expert, after accounting for both load imbalance and grouped-GEMM tile quantization.
When routing statistics are unavailable, EnergyLens can assume uniform routing to support early-stage exploration. When representative workloads are available, these effective values can be computed from a functional model execution to improve accuracy. Importantly, this requires neither an optimized implementation nor distributed profiling: data collected from a single-GPU execution can be transformed into the per-GPU token distribution under expert parallelism.

We extract a max pair $(T_{\max}, E_{\max})$ from the most heavily loaded GPU for latency estimation, and an average pair $(T_{\text{avg}}, E_{\text{avg}})$ across all GPUs for energy estimation.
\Cref{fig:moe_routing} shows example statistics. For prefill, most experts are activated, but $T_{max}$ grows faster than $T_{avg}$ due to load imbalance. In decode, each request produces only one token; the grouped-GEMM kernel enforces a minimum tile granularity of 16 tokens per expert regardless of batch size, though larger batches activate more experts.

Given either ideal or measured routing statistics, per-GPU latency is set by the bottleneck, while per-GPU energy accounts for idle waiting during load imbalance:
\begin{equation}
\mathcal{L} = \sum_i \mathcal{L}_i(T_{\max}, E_{\max}), \qquad
\mathcal{E} = \sum_i \Big[\mathcal{E}_i(T_{\text{avg}}, E_{\text{avg}}) \;+\; (\mathcal{L}_i(T_{\max}, E_{\max}) - \mathcal{L}_i(T_{\text{avg}}, E_{\text{avg}})) \cdot P_{\text{idle}}\Big],
\end{equation}
where $\mathcal{L}_i$ and $\mathcal{E}_i$ are the latency and energy of the $i$-th operation, and $P_{\text{idle}}$ is the GPU idle power.

\subsection{Communication Energy Modeling}
\label{sec:communication}

\begin{wrapfigure}[17]{r}{0.48\columnwidth}
\vspace{-14pt}
    \centering
    \includegraphics[width=0.46\columnwidth]{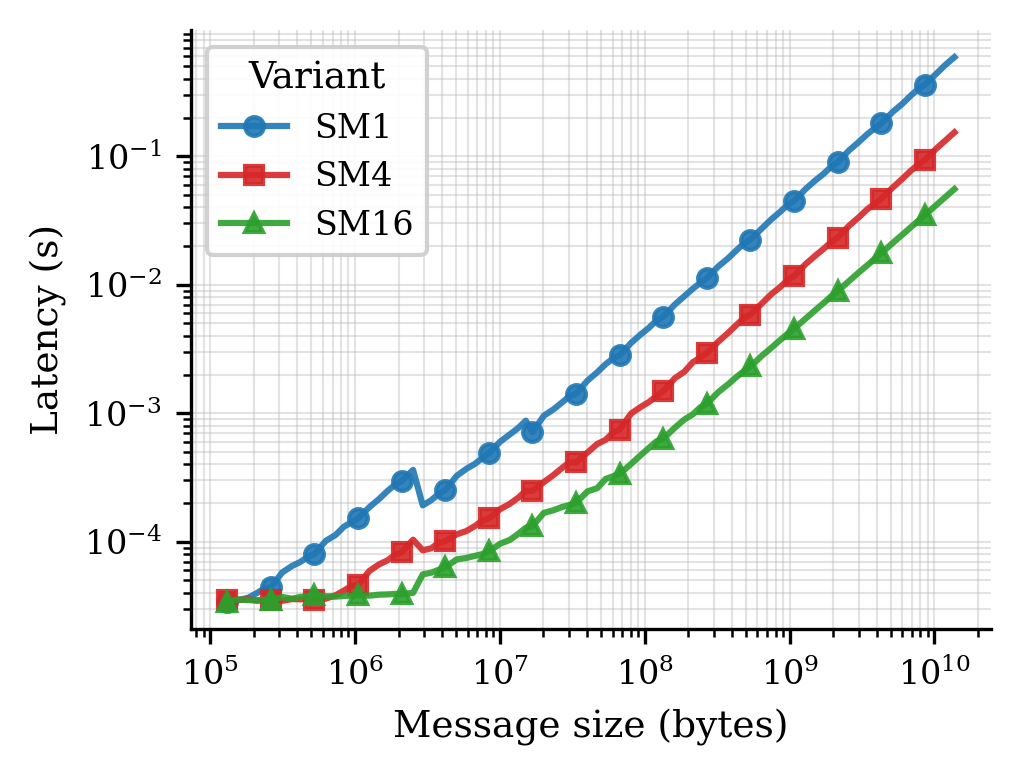}
    \caption{8-GPU ReduceScatter is slower when fewer \acp{SM} are dedicated to communication.}
    \label{fig:RS}
\end{wrapfigure}

We propose an empirically-driven and overlap-aware communication energy model. Unlike prior work that assumes fixed bandwidth utilization and only estimates latency~\citep{neusight}, we profile latency and energy consumption across various communication kernels and transfer sizes and use interpolation during prediction. Overhead dominates at small workload sizes, resulting in higher latency and energy per bit (\Cref{fig:allreduce_energy_latency} in \Cref{sec:app_allreduce}). This behavior is missed by bandwidth-and-queuing-based models used in prior works~\citep{neusight,li2023}. \Cref{fig:RS} further shows that the number of SMs dedicated to communication impacts latency, creating a trade-off between SM allocations for compute and communication during overlap.

\subsection{Compute-Communication Overlap}
\label{sec:overlap}

\begin{figure}[bt]
    \centering
    \begin{subfigure}{0.48\linewidth}
    \centering
    \includegraphics[width=\linewidth]{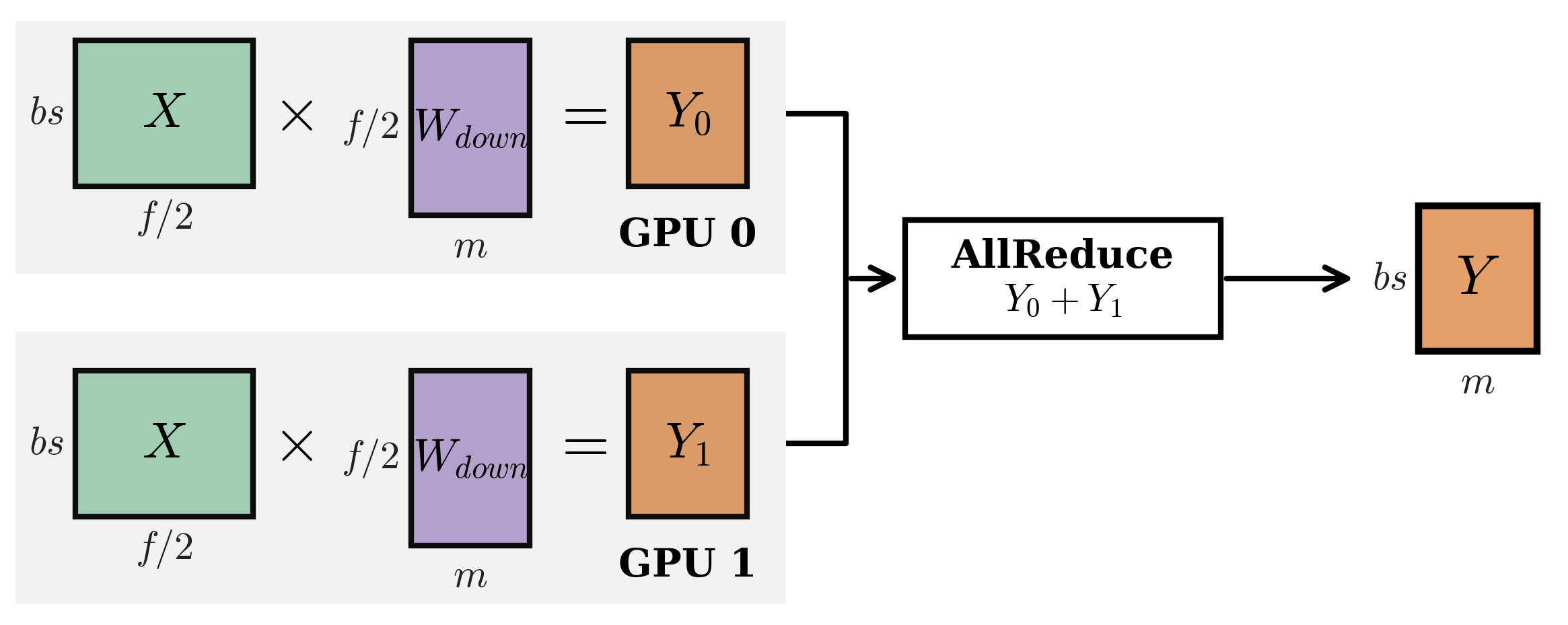}
    \caption{Down projection with TP2.}
    \label{fig:TP_diagram}
    \end{subfigure}
    \hspace{0.02\textwidth}
    \begin{subfigure}{0.48\linewidth}
        \centering
    \includegraphics[width=\linewidth]{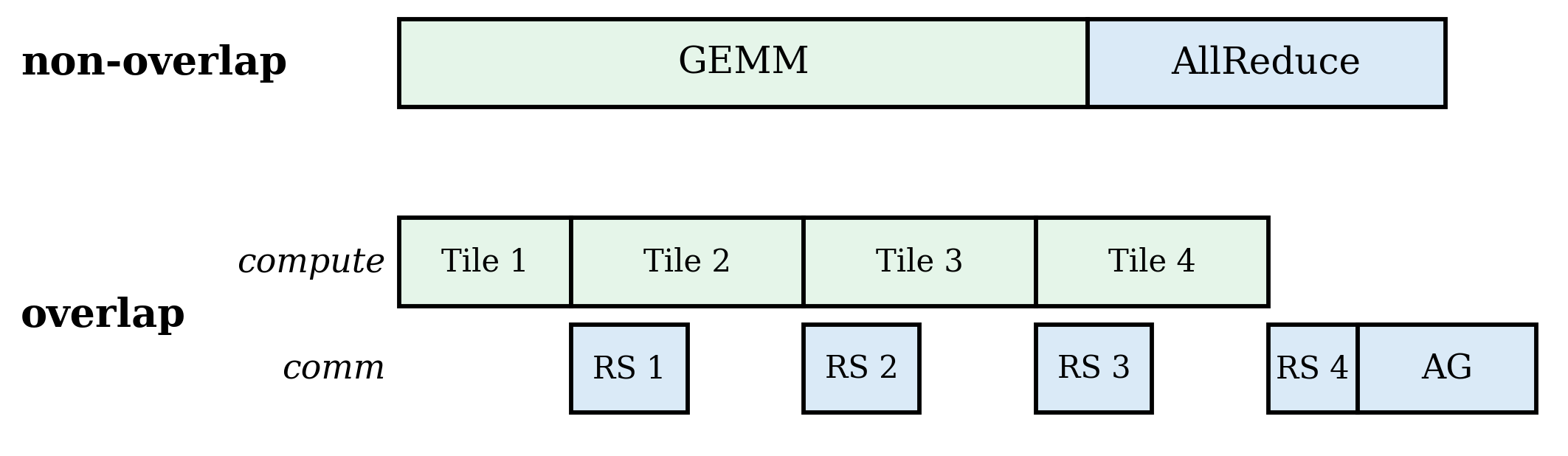}
    \caption{Timeline (RS: ReduceScatter; AG: AllGather).}
    \label{fig:timeline}
    \end{subfigure}
    \caption{\ourwork{} enables intuitive specification of parallelism and overlap settings.}
    \label{fig:TP}
\end{figure}

Distributed inference frameworks increasingly overlap tensor-parallel collectives with GEMMs to hide communication latency~\citep{nemo_overlap}.
EnergyLens models such overlap with three additions to the einsum specification. \texttt{overlap\_stage} sets the number of stages into which compute and communication are partitioned (\Cref{fig:timeline}); \texttt{overlap\_SM} sets the number of SMs dedicated to communication; and \texttt{overlap} specifies the dimension along which this partitioning is applied. For example, a down projection with four-stage overlap and 16 SMs reserved for communication is written as
\lstinline[basicstyle=\ttfamily,keywordstyle={},stringstyle={},identifierstyle={},showstringspaces=false]!op("bsf,fm->bsm", parallel="f", overlap_stage=4, overlap_SM=16, overlap="s", label="Down Projection")!.

EnergyLens models an example GEMM-AllReduce overlapped layer as three phases in \Cref{fig:timeline}: (i) the first non-overlapped compute stage using all SMs, (ii) overlapped stages whose duration is the maximum of the SM-restricted GEMM kernel and the SM-restricted NCCL kernel, and (iii) a final exposed communication stage using all SMs. Stage latencies are obtained by invoking our SM-allocation-aware model for NCCL and using the kernel-level model on the partitioned GEMM. The total latency is then calculated as
\begin{equation}
    t_{total} = t_{GEMM} + t_{RS} 
    + \max\left(t_{GEMM_{overlap}}, t_{RS_{overlap}}\right) \times (stage - 1),
\end{equation}
where $t_{GEMM}$ and related terms denote per-stage latencies.
The latency of an overlapped stage is larger than that of a non-overlapped stage due to fewer SMs being available. The impact of SM allocation on an example communication kernel is shown in \Cref{fig:RS}.
Prior overlap latency model assumes that a quarter-sized partition of GEMM would take a quarter of the latency~\citep{hong2025flux}, which is not accurate at smaller workloads due to decreased algorithmic intensity and kernel overhead. In contrast, EnergyLens queries the compute kernel model for the quarter-sized partition, which is important at small batch-sequence products to accurately determine whether multi-stage overlap is profitable.

For overlapped stages we attribute power using the GEMM estimate, since the majority of SMs and memory bandwidth serve compute. For the exposed communication stage we use the same overlapped-stage power rather than steady-state NCCL power; we find the latter underestimates energy, potentially because GPU DVFS cannot adjust voltage during such short kernel durations.

\subsection{Energy Trade-offs}
We introduce energy-centric metrics analogous to traditional latency metrics for LLM inference: \ac{ETFT} measures the energy consumption during prefill (similar to \ac{TTFT}), while \ac{EPOT} measures decode phase energy normalized by \ac{OSL} (analogous to \ac{TPOT}). Both \ac{ETFT} and \ac{EPOT} are normalized per request.
\ourwork{}'s interface enables users to directly explore these workload-level energy metrics, along with traditional latency and throughput metrics, from a few lines of model descriptions.

\section{Evaluation}
\label{sec:eval}

\subsection{Evaluation Methods}
\label{sec:eval_method}

We evaluate on Llama3-8B, Llama3-70B, and Qwen3-30B-A3B as representative dense and MoE models \citep{llama3_etal, yangQwen3TechnicalReport2025}. Single-GPU evaluation is detailed in \Cref{sec:app_1gpu}, where we validate that high-level einsum specifications are sufficient to predict end-to-end prefill and decode energy before introducing multi-GPU effects. Ground-truth measurements use TensorRT-LLM \citep{nvidia_tensorrt-llm_nodate} on a server with 8 A100-SXM4-80GB GPUs.
To capture steady-state behavior, we apply a 20\,s warmup before measurement. Each workload is then characterized for at least one full iteration and at least 40\,s. We measure end-to-end latency directly and GPU energy via \ac{NVML}.
We use Torch Profiler \citep{pytorch_foundation_torcheinsum_nodate} to validate that the communication collectives are predicted correctly. We also extract the profiled latency breakdown and confirm the actual batch sizes used by TensorRT-LLM, which may differ from the requested values at long contexts. Execution traces also verified that there is no overlap between compute and communication kernels in these experiments.

\input{elements/mape_table}

For compute-communication overlap experiments (\Cref{sec:eval-overlap}), we additionally use Megatron-LM v0.15.3~\citep{shoeybi_megatron-lm_2019} as ground truth. Profiler traces reveal that tensor-parallel all-reduce is decomposed as a reduce-scatter overlapped with GEMM followed by an exposed all-gather.

\subsection{Multi-GPU Inference Evaluation}
\label{sec:multi-gpu}
\begin{figure}[tb]
    \centering
        \begin{subfigure}[b]{0.48\linewidth}
        \centering
    \includegraphics[width=\linewidth]{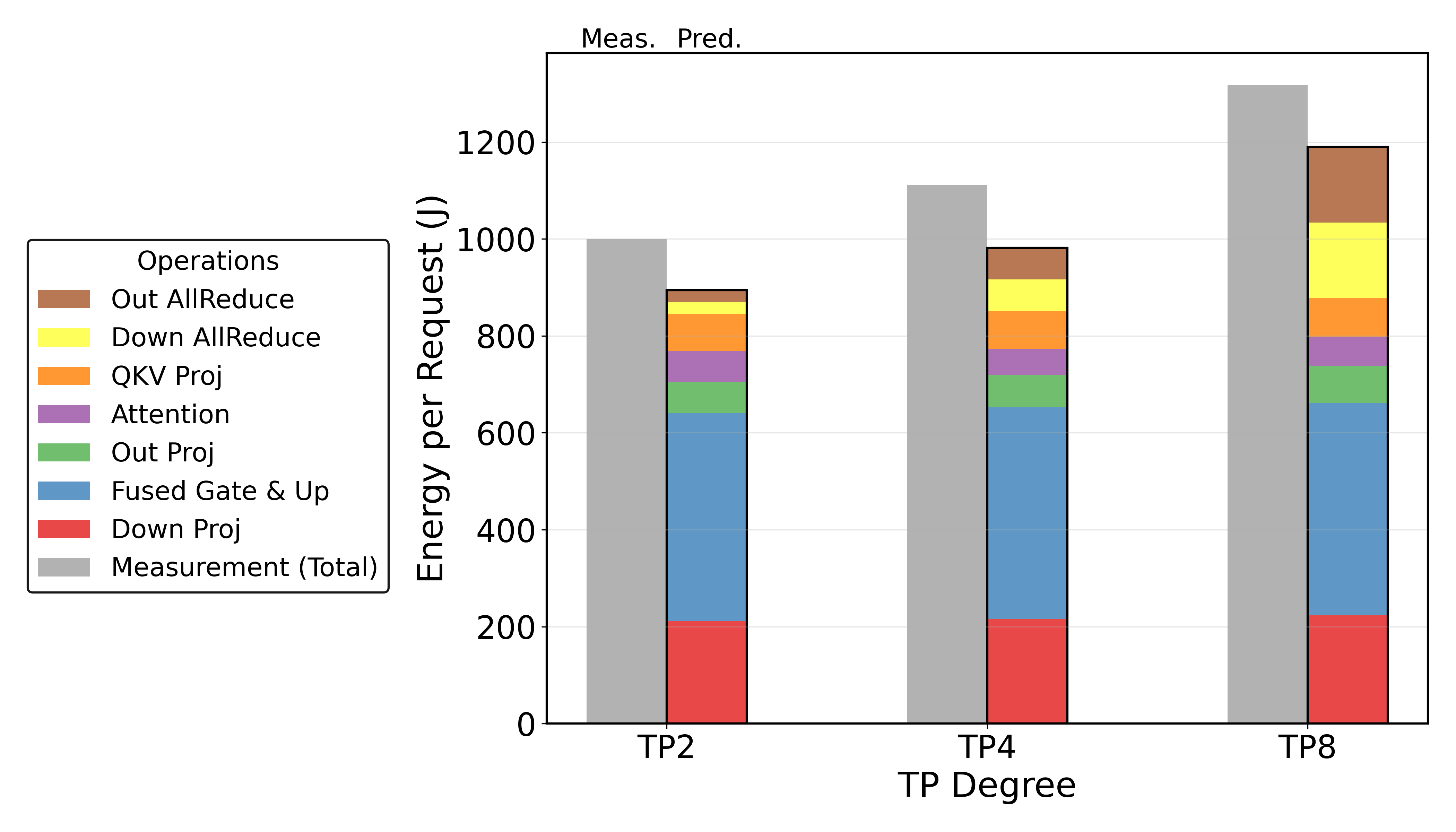}
    \caption{Prefill energy breakdown.}
        \label{fig:70B_prefill_E}
    \end{subfigure}
    \begin{subfigure}[b]{0.48\linewidth}
        \centering
    \includegraphics[width=\linewidth]{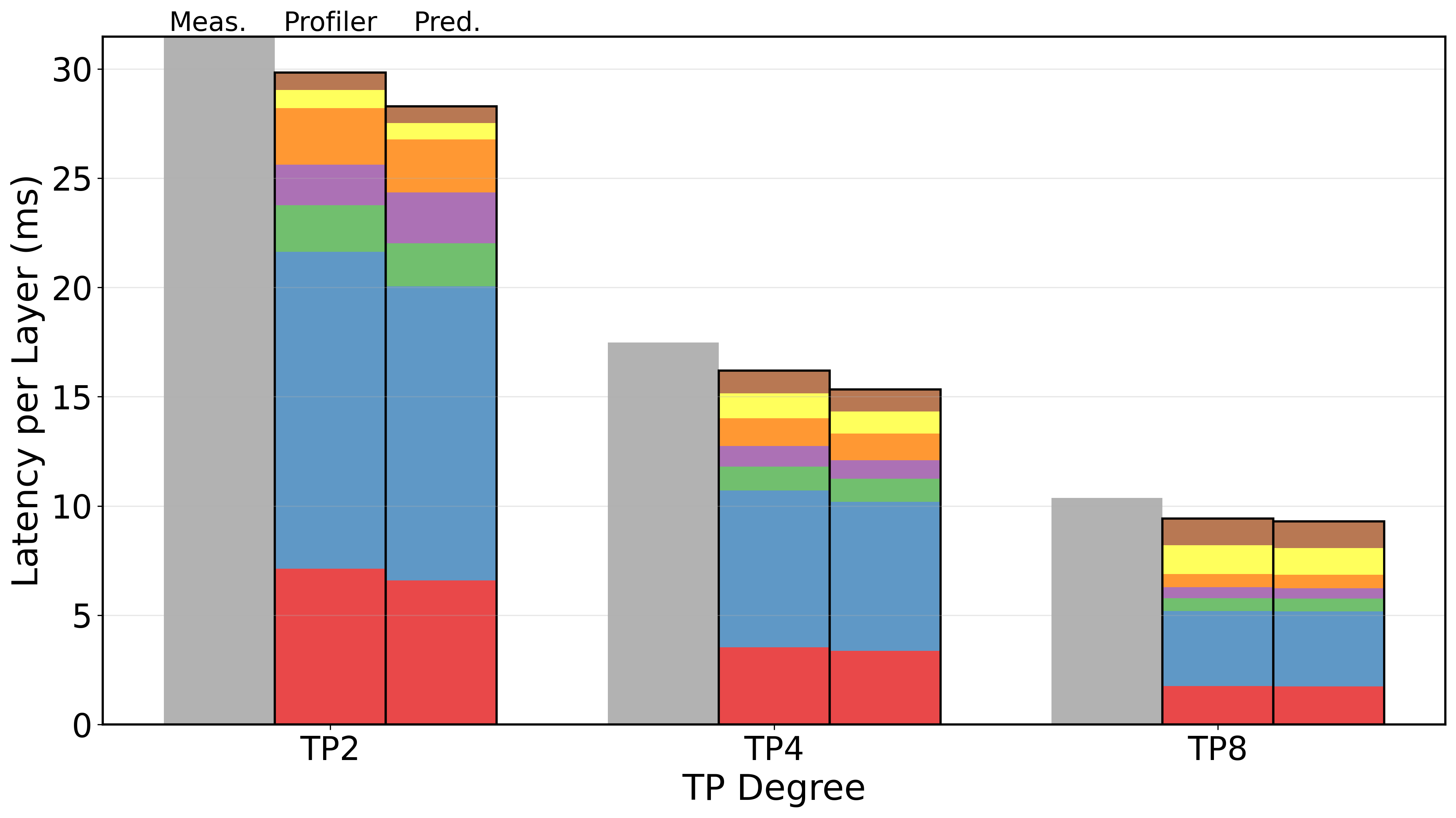}
    \caption{Prefill latency breakdown.}

    \label{fig:70B_prefill_T}
    \end{subfigure}
    \caption{Llama3-70B prefill phase at B2, ISL=4096 across tensor parallelism configurations (TP2, TP4, TP8), showing increasing communication overhead as TP degree increases. The grey bars are the measurement, and the rightmost bars in each cluster are the \ourwork{} prediction.}
    \label{fig:70B_prefill}

\end{figure}

We report the overall prediction errors on dense and MoE models with various parallelism configurations in \Cref{tab:mape} (sweep parameters specified in \Cref{sec:app_setup}).
For prefill, most operators are compute-bound at practical \acp{ISL}, so further increasing arithmetic intensity through batching yields less energy efficiency gain than lower tensor parallelism. \Cref{fig:70B_prefill_E} shows that compute energy remains nearly constant across \ac{TP} configurations, while AllReduce cost grows with TP. At TP8, AllReduce accounts for 23\% of total energy.
\Cref{fig:70B_prefill_T} further shows that the predicted latency breakdown closely matches the measured execution trace, validating the kernel-level decomposition used for energy attribution.
We observe the same trend on Qwen3-MoE (\Cref{fig:qwen_prefill} in \Cref{sec:app_qwen_prefill}): compute energy is stable across \ac{TP} degrees while communication overhead grows.

\begin{figure}[tb]
    \centering
        \begin{subfigure}[b]{0.48\linewidth}
        \centering
    \includegraphics[width=\linewidth]{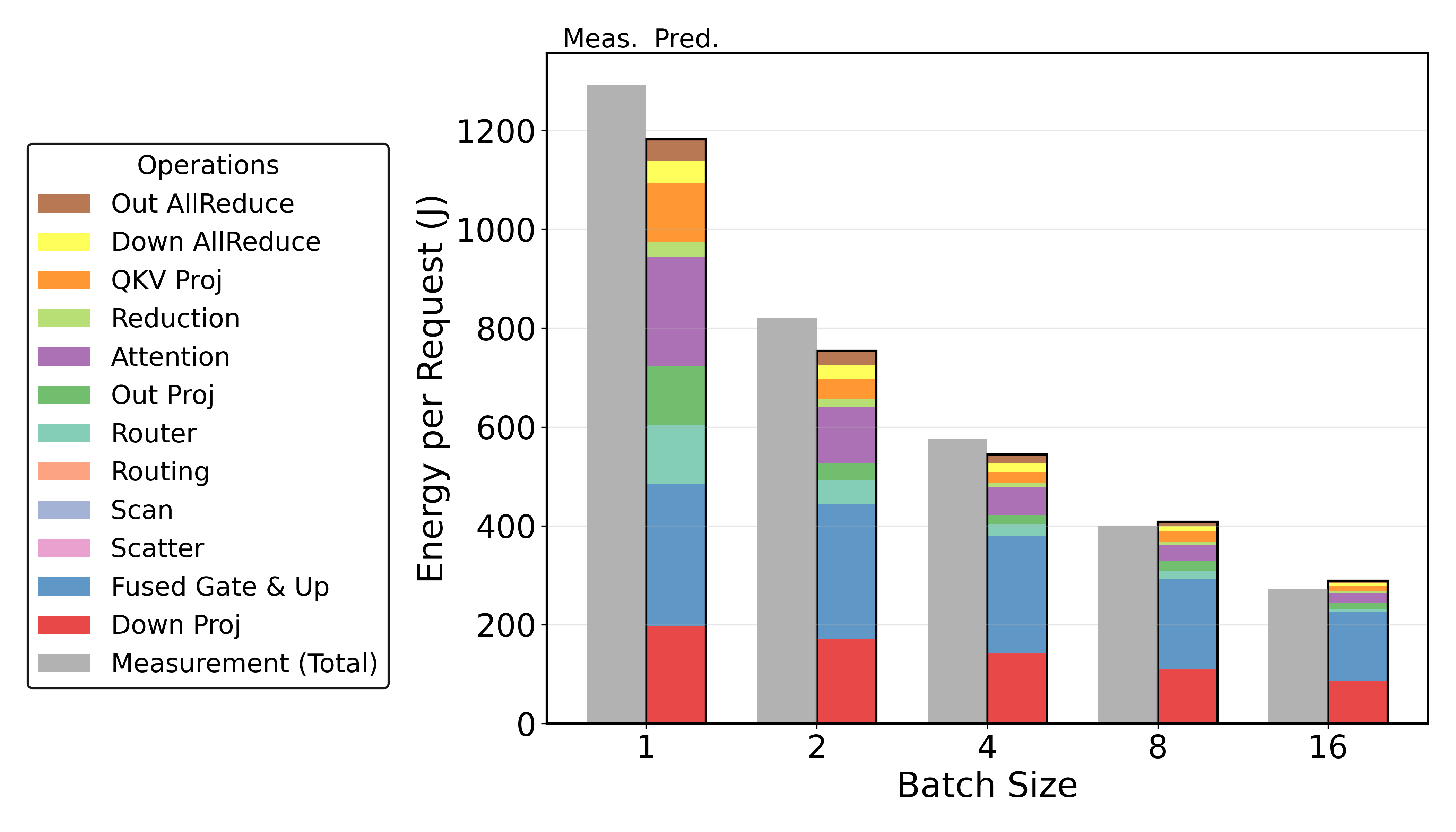}
    \caption{Decode energy breakdown.}
        \label{fig:qwen_decode_E}
    \end{subfigure}
    \begin{subfigure}[b]{0.48\linewidth}
        \centering
    \includegraphics[width=\linewidth]{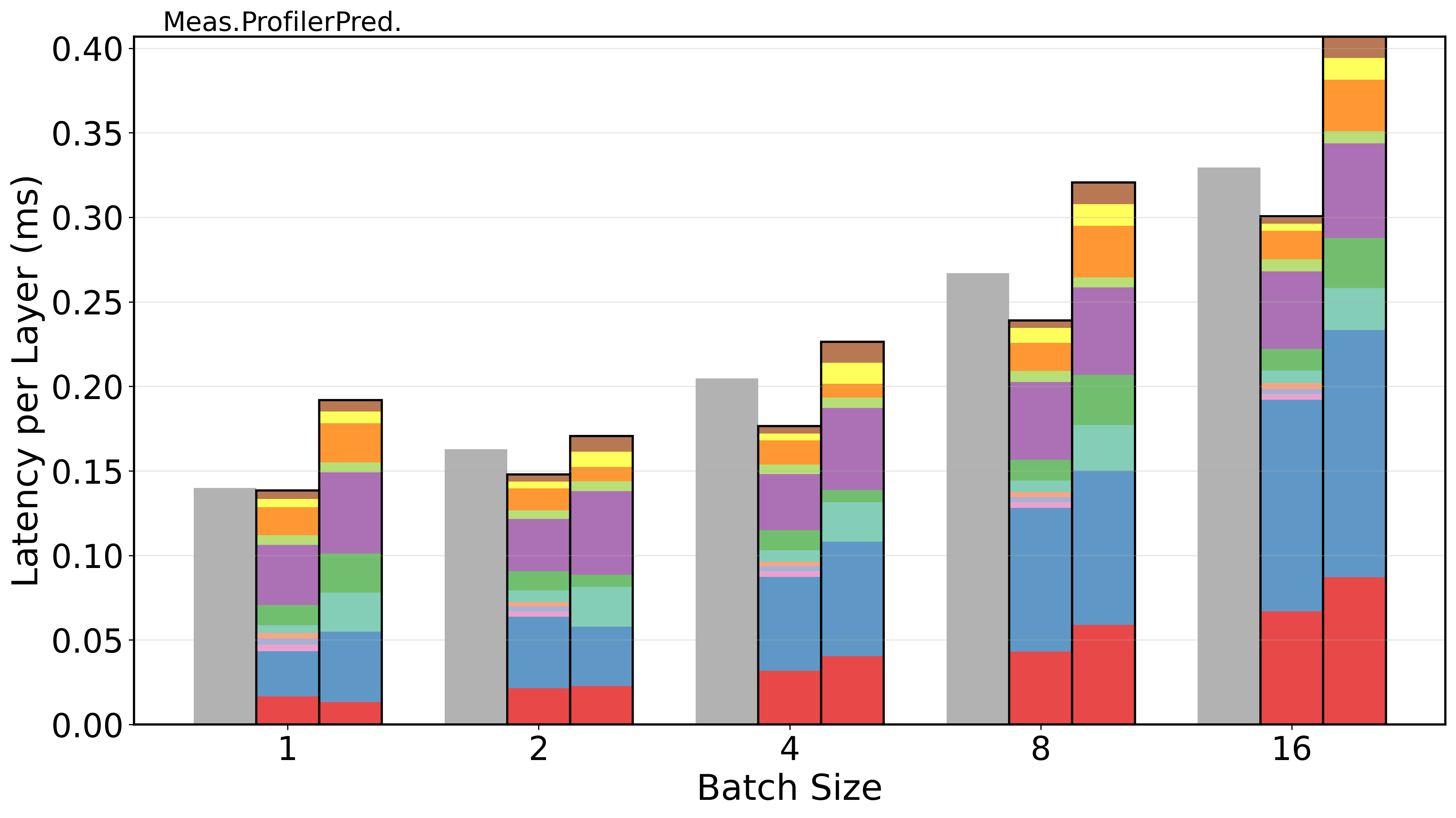}
    \caption{Decode latency breakdown.}
    \label{fig:qwen_decode_T}
    \end{subfigure}
    \caption{Qwen3-30B-A3B decode phase at TP2 in attention layers and EP2 in MoE layers, with ISL=512 and OSL=1024. Larger batches substantially reduce decode energy per token.}
    \label{fig:qwen_decode}

\end{figure}

Decode exhibits a different trade-off, where energy varies more strongly with batch size than with \ac{TP}, and larger batches substantially reduce decode energy. However, batching is constrained by GPU memory, since increasing batch size reduces the maximum sequence length that can fit in DRAM. This trade-off arises because decode processes 1 token per step, producing skewed GEMMs with low arithmetic intensity. In MoE models, this effect manifests differently across operation types. The output projections behave similarly to dense models: for example, the output projection \texttt{op("bsHh,Hhm->bsm", parallel="H")} maps to a matmul with $M{=}b$ and $N{=}m$.
At small batch sizes, these kernels are memory-bound and underutilize the GPU, while scaling up batch sizes increases arithmetic intensity and reduces energy per output token.
The expert layers depend on the effective tokens per expert, $T$, and activated experts per GPU, $E$. The gate and up projection \texttt{op("ETm,EmF->ETF", parallel="E")} maps to a grouped GEMM with $M{=}T$ and $N{=}F$, where minimum tile size forces $T$ to be at least 16 regardless of batch size during decode (\Cref{fig:moe_routing}). This means that at small batch sizes, these GEMMs are also padded and severely underutilized. While larger batches activate more experts, it also increases the number of useful tokens per expert, thereby improving utilization.

Decode latency is harder to predict accurately, because compute backends behave worse with low-arithmetic-intensity kernels and highly skewed matrix shapes present in the decode phase of both standard and MoE models. We observe similar limitations across alternative compute backends \citep{neusight,li2023}, indicating that this issue is not backend-specific (details in \Cref{sec:limit}). Nevertheless, \ourwork{} still captures the key trend relevant for design exploration: decode latency changes only modestly with batch size, while energy improves substantially.

\subsection{Compute-Communication Overlap}
\label{sec:eval-overlap}

We validate overlap modeling on Llama3-70B with overlap configurations
including no overlap and 4-stage overlap with 1, 4, and 16 SMs
dedicated to communication (full sweep parameters specified in \Cref{sec:app_setup}).
Across this sweep, EnergyLens predicts energy with a MAPE of
\overlapE\ and latency with a MAPE of \overlapT.
More importantly, EnergyLens helps to predict the relative
trade-offs across SM allocations: dedicating too many SMs to
communication wastes compute resources, while allocating too few makes communication the bottleneck of the overlapped stages. The ability to include overlap as a dimension in energy-latency trade-off will be exploited in \Cref{sec:dse_overlap}.

\section{Energy-Centric Exploration}

The results in \Cref{sec:eval} show that EnergyLens’s energy prediction error is small relative to the variations across practical deployment choices. What matters most for optimization is preserving the ordering between candidate configurations to identify favorable regions of the design space rather than exact pointwise estimation. In this section, we further show that EnergyLens' accuracy enables us to identify optimal configurations, not merely perform post-hoc energy analysis. We leverage this property below to derive energy-driven insights on batching, tensor parallelism, disaggregated serving, and overlap tuning.

\subsection{Disaggregated Serving}
\label{sec:dse}

\begin{figure}[!htbp]
    \centering
    \begin{subfigure}[b]{0.48\linewidth}
        \centering
        \includegraphics[width=\linewidth]{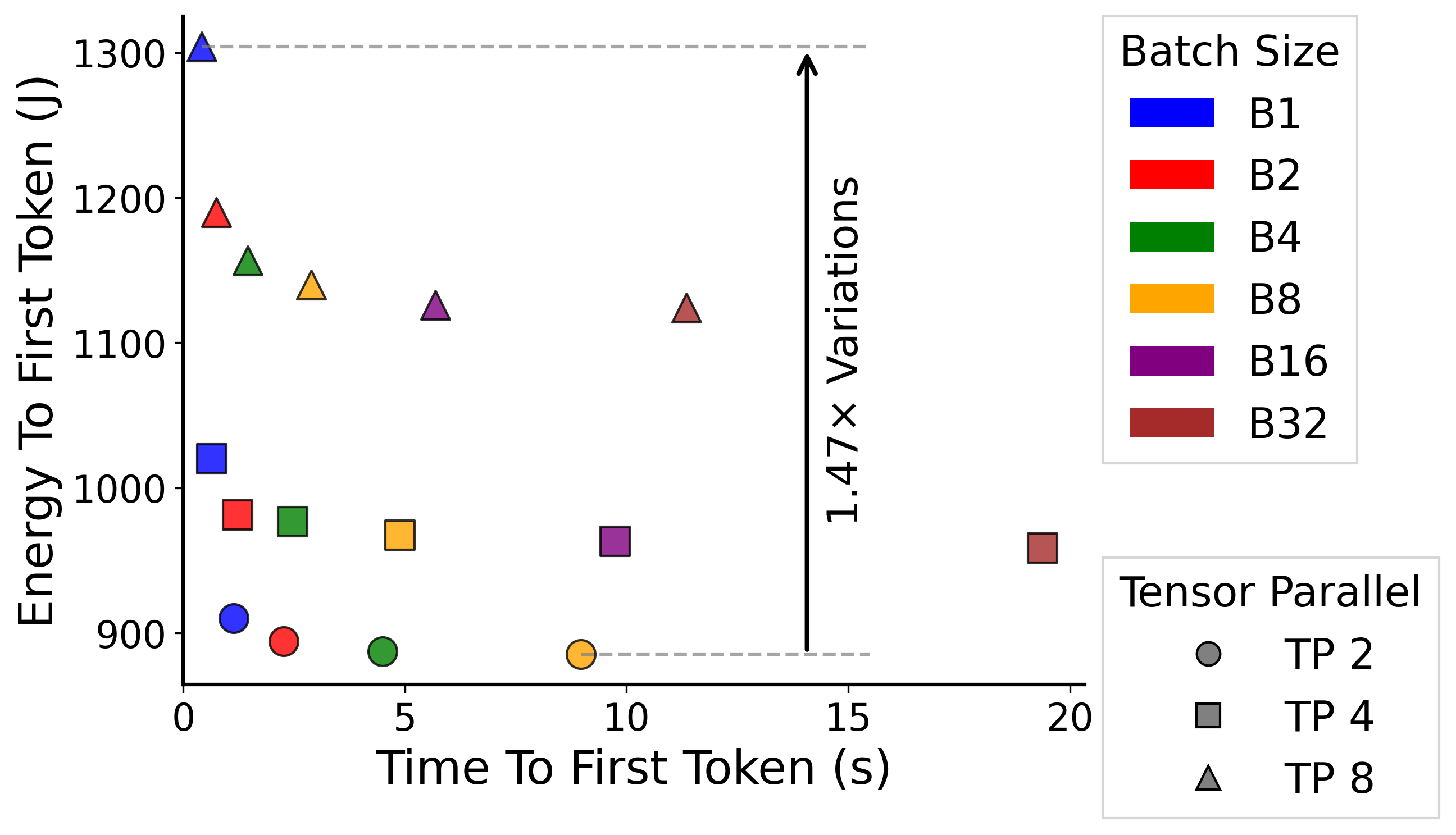}
        \caption{Prefill (ISL=4096): TP2 dominates the Pareto frontier due to reduced communication overhead.}
        \label{fig:dse_prefill}
    \end{subfigure}
    \hfill
    \begin{subfigure}[b]{0.48\linewidth}
        \centering
        \includegraphics[width=\linewidth]{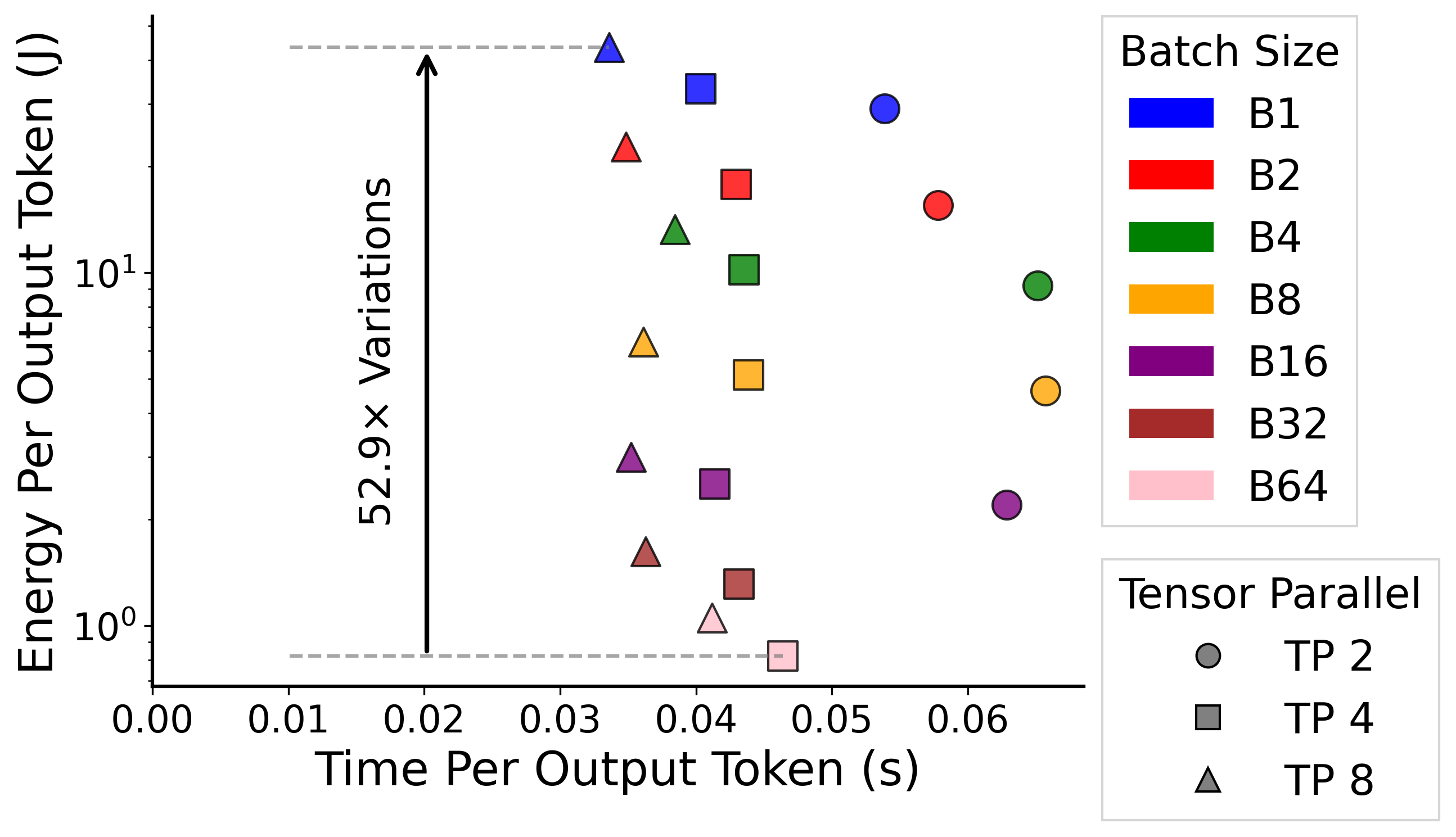}
        \caption{Decode (ISL=512, OSL=1024): larger batch sizes achieve order-of-magnitude \ac{EPOT} improvement.}
        \label{fig:dse_decode}
    \end{subfigure}
    \caption{Energy-latency trade-offs for Llama3-70B. Configurations exceeding GPU memory are omitted.}
    \label{fig:dse_prefill_decode}
\end{figure}

\textit{\textbf{Key Insight 1:} At equal \ac{TTFT}, smaller batch size and lower tensor parallelism is more energy efficient.}
Two approaches can reduce \ac{TTFT}: smaller batch sizes decrease compute per batch, while higher \ac{TP} distributes the same compute across more GPUs.
\Cref{fig:dse_prefill} reveals that most Pareto-optimal configurations use TP2, due to reduced communication overhead (\Cref{fig:70B_prefill_E}); the lower bound on TP degree comes from GPU memory requirements.
At \ac{ISL} 4096, TP2 with batch size 4 is both faster and consumes 24\% less energy than TP8 with batch size 16. This is confirmed by measurements, which show 30\% energy savings. Across all ISLs for 70B prefill, B4 TP2 over B16 TP8 yields an average \prefillred{} energy reduction with lower latency.

\textit{\textbf{Key Insight 2}: Increasing batch size yields significant energy savings during decode, at minimal latency cost.}
\Cref{fig:dse_decode} plots \ac{EPOT} against \ac{TPOT}: larger batch sizes dramatically reduce \ac{EPOT} with only modest latency increase. At batch size 1, skewed matrix shapes ($M = b = 1$) severely underutilize GPUs; increasing batch size shifts \ac{GEMM} kernels toward compute-bound regimes, improving efficiency. Across \ac{OSL} up to 7680 and TP8, batch size 16 achieves \decodered{} \ac{EPOT} reduction versus batch size 1. The memory requirement of larger batch size can be mitigated with higher degrees of parallelism.
These contrasting objectives, lower batch sizes for prefill latency versus larger batch sizes for decode efficiency, strongly motivate disaggregated serving~\citep{distserve}, where prefill and decode are independently optimized.

\subsection{Megatron-Style Overlap}
\label{sec:dse_overlap}

\textit{\textbf{Key Insight 3:} Overlap tuning is hard to reason with intuition alone, and EnergyLens identifies Pareto-optimal configurations much more reliably than a naive maximum-overlap approach.}

\begin{wrapfigure}[22]{r}{0.48\columnwidth}
\vspace{-14pt}
    \centering
    \includegraphics[width=0.46\columnwidth]{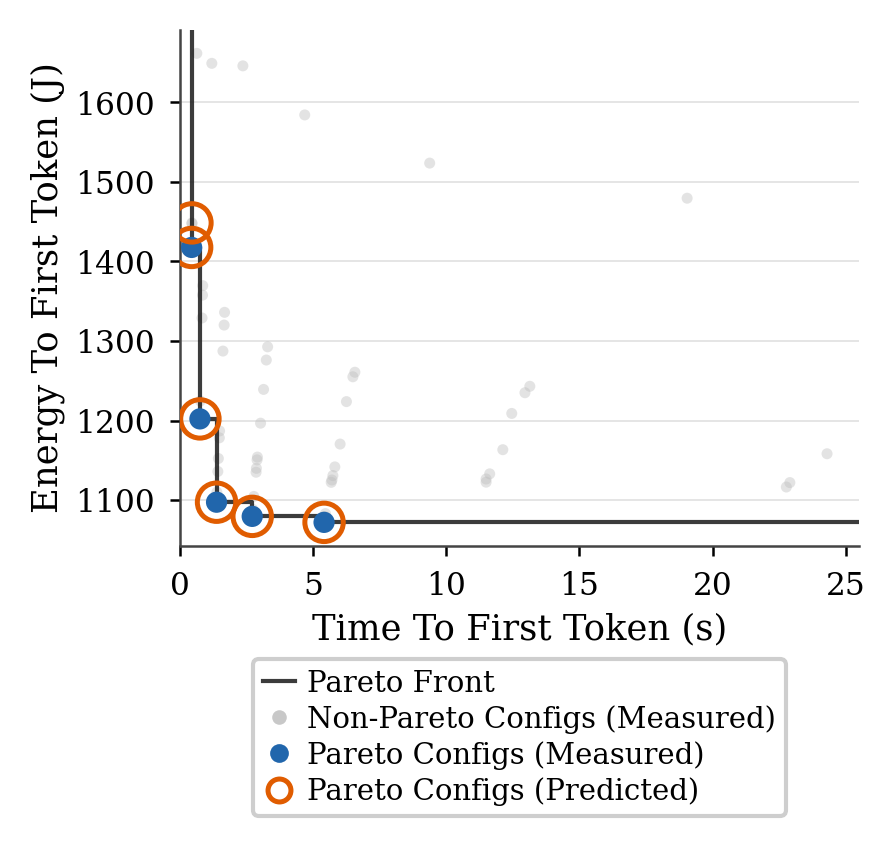}
    \caption{Llama3-70B (ISL=4096): the predicted Pareto-optimal points fall on the Pareto Front of 55 measured configurations with varying overlap settings, batch size, and tensor parallelism.}
    \label{fig:megatron_meas}
\end{wrapfigure}

Overlap introduces another optimization dimension beyond tensor parallelism and batch size. Instead of relying on intuition to prioritize maximum overlap, EnergyLens predicts the full configuration space and identifies the Pareto-optimal configurations. \Cref{fig:megatron_meas} shows this at ISL 4096 across TP degrees, batch sizes, and communication SM allocations. In this example, the Pareto frontier is dominated by non-overlapped TP2 settings, with overlap becoming beneficial only at the most latency-aggressive point. This illustrates why overlap is difficult to tune heuristically: some Pareto points avoid overlap altogether and reduce communication through lower TP.

We measure the exact Pareto frontiers in 32 scenarios with different \acp{ISL}, allowing both overlapped and non-overlapped settings. A common intuition is to use maximum overlap (16 SMs for communication) and sweep only batch size and tensor parallelism. This maximum-overlap heuristic recovers only \OverlapExactRecall{} of the exact measured Pareto frontier, because it cannot discover configurations where disabling overlap or dedicating a moderate number of SMs to communication is preferable. In contrast, EnergyLens predicts energy and latency for all 55 configurations, selects the predicted Pareto-optimal subset, which recovers \FullExactRecall{} of the exact frontier. This is achieved without any measurement.
\Cref{fig:megatron_meas} confirms that the configurations selected by EnergyLens lie on the measured Pareto frontier. This shows that the value of EnergyLens is not merely in shrinking the search space, but in directly identifying the Pareto configurations without measurement.

\section{Conclusion}

We present \ourwork{}, a framework for predicting LLM inference energy from a high-level specification without requiring implementations, traces, or GPU access. \ourwork{} models dense and MoE models, multi-GPU communication, and Megatron-style overlap, achieving multi-GPU energy \acfp{MAPE} between \qwenTEPE{} and \qwenTEPdecodeE{}, sufficient for identifying optimal deployment regimes.

Despite the limitation in lower decode latency accuracy for small workloads, \ourwork{} prediction is effective for energy-aware design-space exploration: \ourwork{} reveals large energy variation across configurations, motivates disaggregated serving, and identifies Pareto overlap settings significantly better than the naive maximum-overlap approach.

%% file: elements/llm_spec_qwen.tex
\begin{listing}[th]
\begin{minted}[
    fontsize=\small,
    linenos,
    xleftmargin=1em,
    breaklines,
    breaksymbolright=,
    breaksymbolleft=\hspace{2em},
    escapeinside=||
]{python}
llm_MoE.EQS = [
    op("bsm,miKh->bsiKh", parallel="K", label="QKV Projection"),|\label{line:qkv}|
    op("attention", parallel="H", attn_eqs=attn_eqs, label="Attention"),|\label{line:attn}|
    op("bsHh,Hhm->bsm", parallel="H", label="Output Projection"),|\label{line:out}|
    op("bsm,Em->bsE", label="Router"),|\label{line:router}|
    op("bsm->bsmA", label="Scatter"),|\label{line:scatter}|
    op("ETm,EmF->ETF", parallel="E", label="Gate & Up Projection"),|\label{line:gateup}|
    op("ETf,Efm->ETm", parallel="E", label="Down Projection"),|\label{line:down}|
    op("bsAm,bsA->bsm", parallel="A", label="Reduction")|\label{line:reduce}|
]
\end{minted}
\caption{Full specification of a fused MoE implementation with \acs{TP} in the attention block and \acs{EP} in the MoE block (symbols and \texttt{attn\_eqs} defined in \Cref{sec:app_spec}).}
\label{lst:qwen}
\end{listing}

%% file: elements/mape_table.tex
\begin{table}[tbhp]
\centering
\small
\caption{\Acfp{MAPE} on dense and MoE models are sufficient for design space optimization.}
\label{tab:mape}
\begin{tabular}{ll ccc cc}
\toprule
& & \textbf{Llama3-8B} & \multicolumn{2}{c}{\textbf{Llama3-70B}} & \multicolumn{2}{c}{\textbf{Qwen3-30B-A3B}} \\
\cmidrule(lr){3-3} \cmidrule(lr){4-5} \cmidrule(lr){6-7}
& & Single GPU & \Ac{TP} & \Ac{TP} (Overlap) & Single GPU & \ac{TP}+\ac{EP} \\
\midrule
\multirow{2}{*}{Prefill} & Energy  & \prefillsingleE & \prefillmultiE & \overlapE & \qwenPrefillE & \qwenTEPE \\
                          & Latency & \prefillsingleT & \prefillmultiT & \overlapT & \qwenPrefillT & \qwenTEPT \\
\addlinespace
\multirow{2}{*}{Decode}  & Energy  & \decodesingleE & \decodemultiE & N/A\textsuperscript{\dag} & \qwenDecodeE & \qwenTEPdecodeE \\
                          & Latency & \decodesingleT & \decodemultiT & N/A\textsuperscript{\dag} & \qwenDecodeT & \qwenTEPdecodeT \\
\bottomrule
\multicolumn{7}{c}{\footnotesize \textsuperscript{\dag}Multi-stage compute-communication overlap is a prefill technique and does not apply to decode} \\
\multicolumn{7}{c}{\footnotesize because the smaller partition would further reduce the compute intensity of the GEMM operation.} \\
\end{tabular}
\end{table}

%% file: appendices.tex
\clearpage
\newpage

\section{Motivation - Power Measurements}
\begin{figure}[h]
    \centering
    \includegraphics[width=0.6\linewidth]{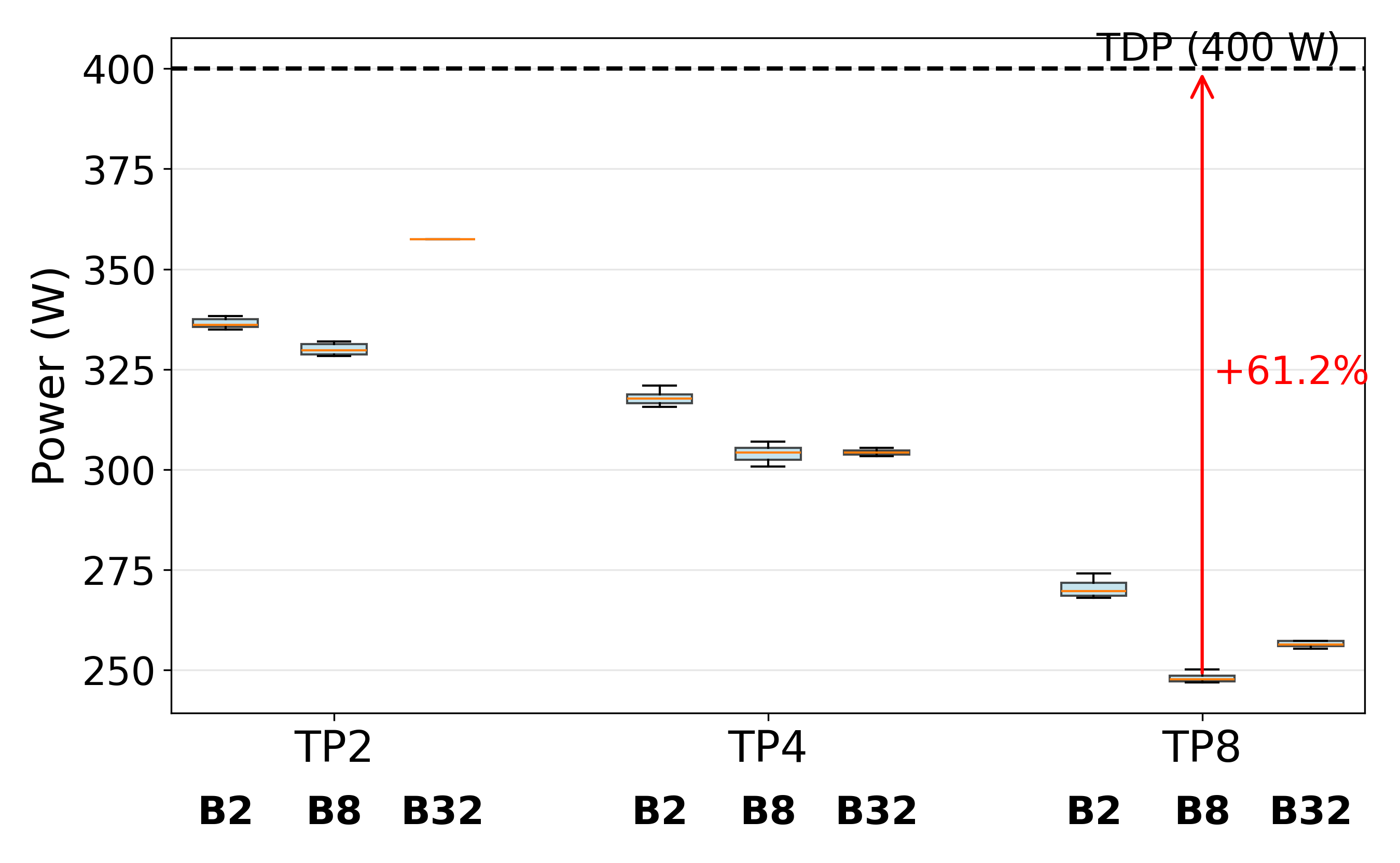}
    \caption{Observed GPU power consumption varies by up to 60\% in the decode phase, not captured by \ac{TDP}. Llama3-70B inference on different numbers of A100-80GB GPUs (TP2, TP4, and TP8) and batch sizes (B2, B8 and B32).}

    \label{fig:motivation_power}
\end{figure}

Latency multiplied by \ac{TDP} does not accurately capture the energy consumption, as power varies greatly especially during the decode phase.

\clearpage
\newpage
\section{Model Specification}
\label{sec:app_spec}
\Cref{lst:ops} is used in both Llama3-8B and Llama3-70B evaluations in \Cref{sec:app_1gpu} and \Cref{sec:multi-gpu}. It corresponds to the fusion strategy used in TensorRT-LLM verified via Torch traces.
\input{elements/llm_spec_unfused}
\input{elements/llm_spec_fused}
\input{elements/attn_eq}
\input{elements/llm_sym}

\clearpage
\newpage
\section{Communication Kernel Detection}
\label{sec:get_comm}
\input{elements/allreduce}
\input{elements/all2all}
\Cref{alg:all2all,alg:allreduce} show the non-overlapped kernel detection mechanism. Compute-communication overlap modeling is detailed in \Cref{sec:overlap}.

\clearpage
\newpage
\section{Additional Plots for Communication}
\label{sec:app_allreduce}

\begin{figure}[h]
    \centering
    \begin{subfigure}[b]{0.48\linewidth}
        \includegraphics[width=\linewidth]{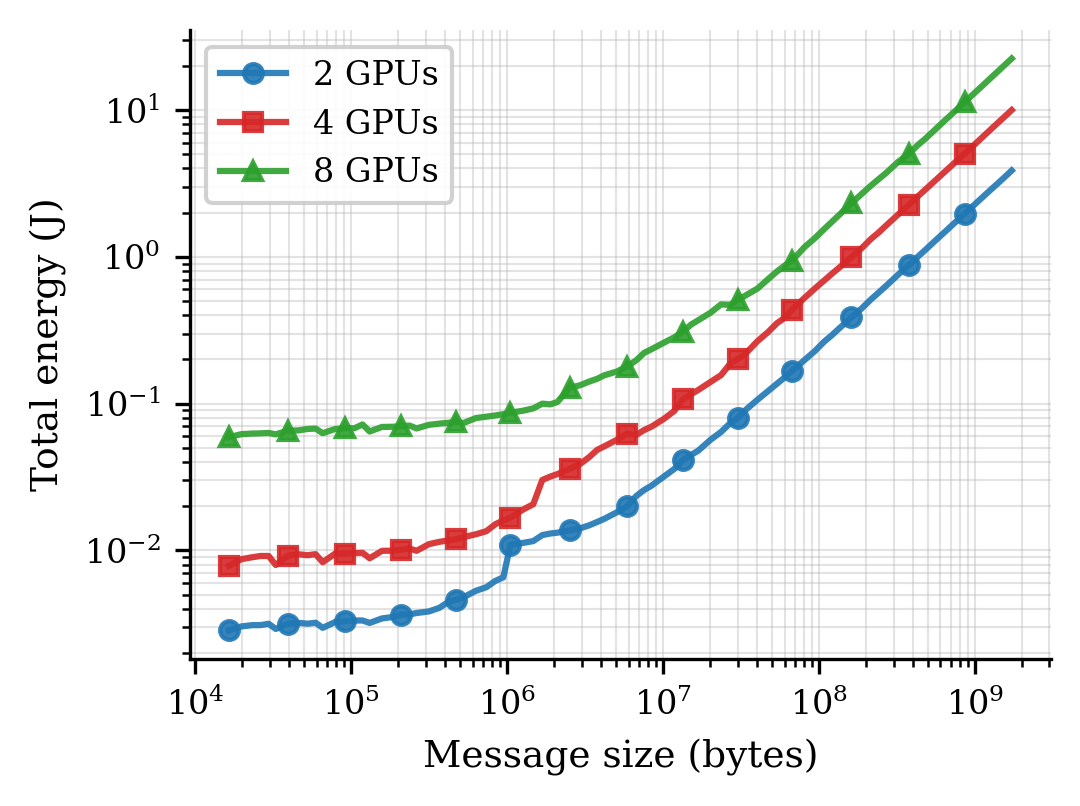}
        \caption{Energy}
        \label{fig:total_energy_overlay}
    \end{subfigure}
    \hfill
    \begin{subfigure}[b]{0.48\linewidth}
        \includegraphics[width=\linewidth]{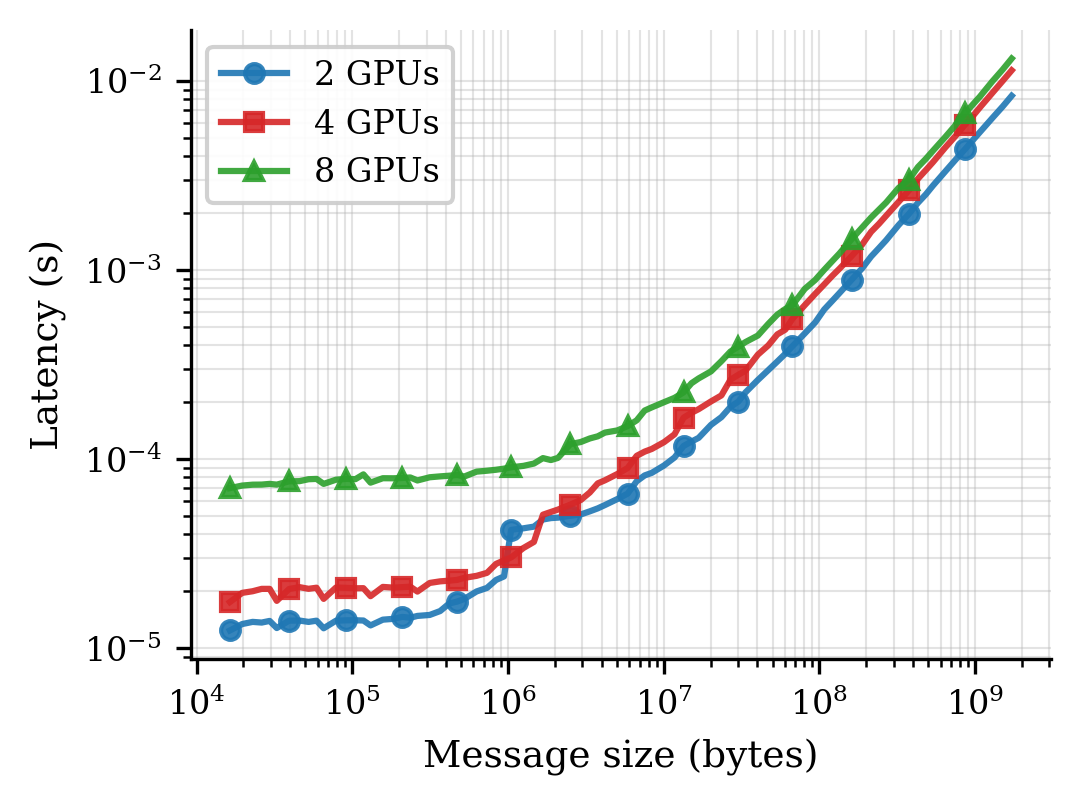}
        \caption{Latency}
        \label{fig:latency_overlay}
    \end{subfigure}
    \caption{Total energy and latency of \texttt{AllReduce} for 2, 4, and 8 GPUs.}
    \label{fig:allreduce_energy_latency}
\end{figure}

Overhead dominates at small workload sizes, resulting in higher latency and energy per bit. This behavior is missed by bandwidth-and-queuing-based models used in prior works~\citep{neusight,li2023}.

\clearpage
\newpage
\section{Single-GPU Evaluation}
\label{sec:app_1gpu}
\begin{figure}[h]
    \centering
    \includegraphics[width=0.6\linewidth]{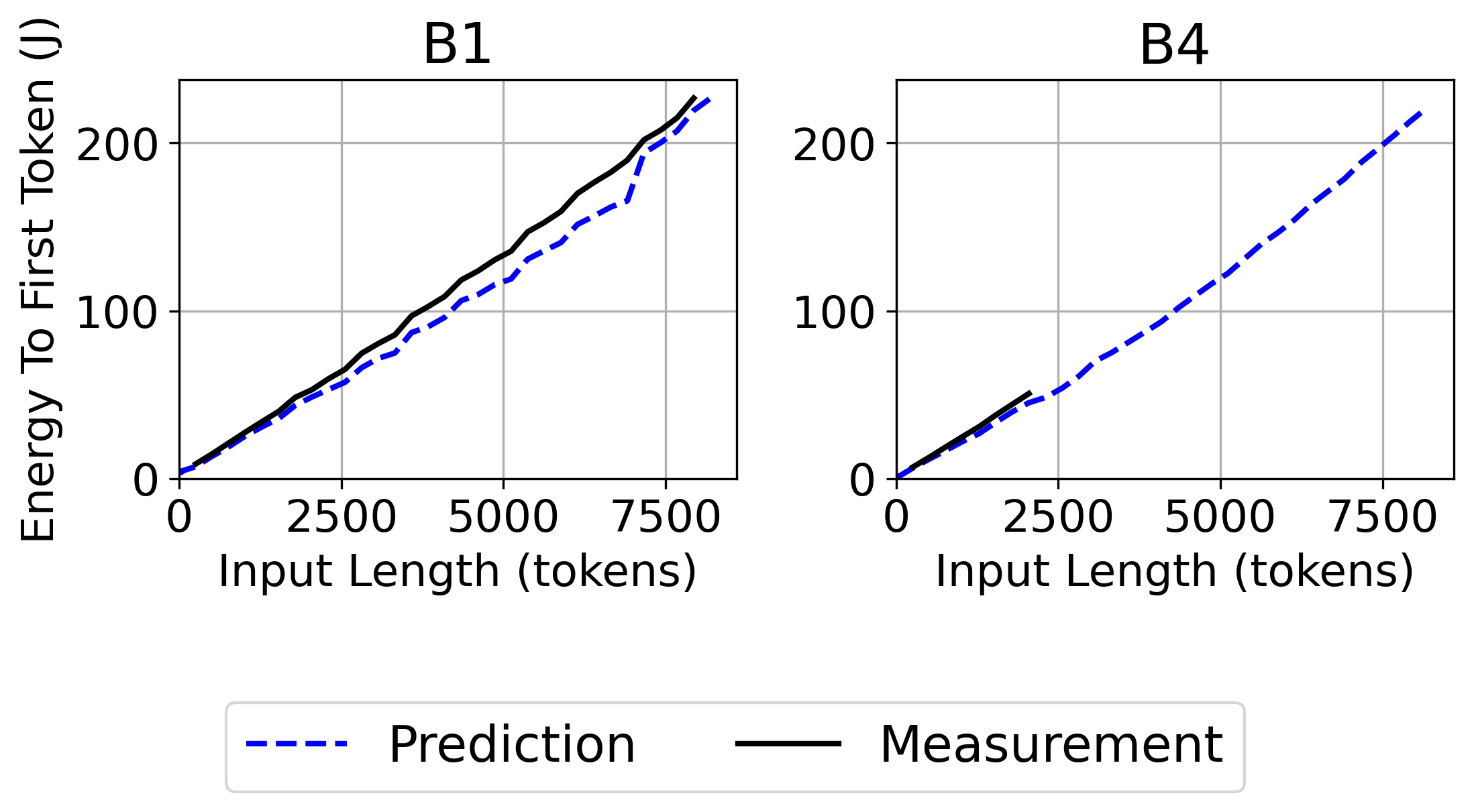}
    \caption{The energy to first token (ETFT) prediction is validated on single-GPU Llama3-8B inferences with a MAPE of \prefillsingleE{}. The B4 measurement curve ends at 2048 input tokens where TensorRT-LLM switches to smaller batch sizes.}

    \label{fig:8B_prefill}
\end{figure}

\begin{figure}[htbp]
    \centering
    \includegraphics[width=0.6\linewidth]{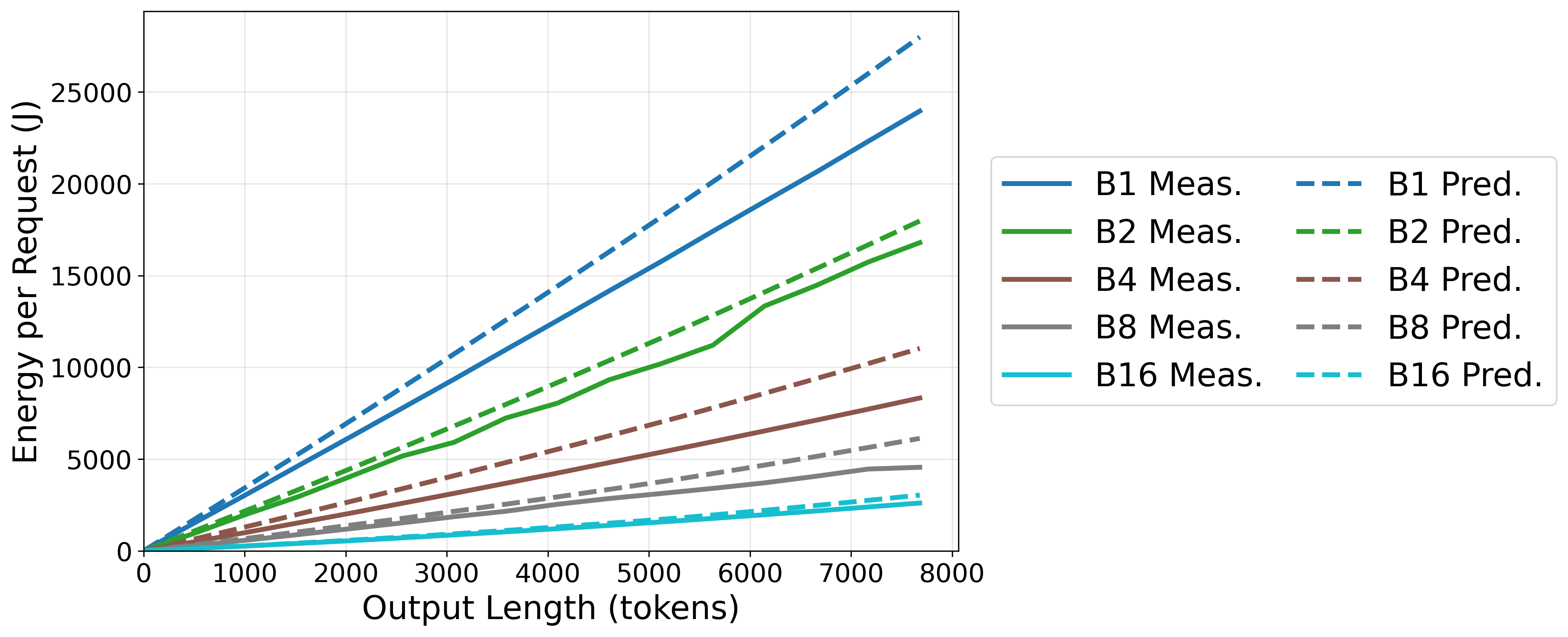}
    \caption{The energy per request prediction is validated on single-GPU Llama3-8B inferences with a MAPE of \decodesingleE{}.}
    \label{fig:8B_decode}
\end{figure}

We first validate \ourwork{} in the single-GPU setting with Llama3-8B. This evaluation tests whether our high-level einsum-based specification is sufficient to predict end-to-end inference energy from model structure alone, without requiring implementation-level kernel details from the user. It also provides a clean evaluation before introducing multi-GPU communication effects.

For prefill, all input tokens are processed simultaneously by the LLM. We sweep \ac{ISL} linearly up to 8192 tokens and batch size logarithmically up to 64. Because TensorRT-LLM may change the realized batch size at long contexts, we use Torch traces to verify the actual executed batch size: for example, at 7936 \ac{ISL}, the engine always uses batch size 1. We include all observed \ac{ISL}-batch-size pairs in the \ac{MAPE} calculation. \Cref{fig:8B_prefill} plots the predicted \ac{ETFT}, normalized per request. Since these operations already have high arithmetic intensity, \ac{ETFT} is largely insensitive to batch size. \ourwork{} closely matches measurements, achieving a MAPE of \prefillsingleE{}.

Decode behaves very differently from prefill. Since only the most recent token is processed at each iteration ($s=1$), decode kernels have lower arithmetic intensity and lower GPU utilization. While prior work validates prefill latency or energy \citep{neusight,energaizer}, decode is often overlooked. We therefore validate decode predictions in \Cref{fig:8B_decode}. Comparing the y-axes of \Cref{fig:8B_prefill,fig:8B_decode}, decoding 4000 output tokens consumes orders of magnitude more energy than prefilling 4000 input tokens. Most importantly, \ourwork{} correctly predicts the key trend relevant for optimization: at fixed \ac{OSL}, energy per request decreases substantially with larger batch sizes, in contrast to prefill behavior.

\clearpage
\newpage
\section{Evaluation for Llama3-8B with \ac{CP}}
\label{sec:app_llama_cp}
\Acf{CP} is increasingly used for long context prefill \citep{jacobs_DeepSpeedUlyssesSystem_}. \ourwork{} inserts the appropriate communication kernels as described in \Cref{alg:all2all}. An example specification of the fused dense transformer with \ac{CP} is provided below.

We validated \ourwork{}'s support for context parallelism (the variety proposed by DeepSpeed-Ulysses) on Llama3-8B. This is tested on CP2 with the same sweep settings described in \Cref{sec:app_setup}, achieving MAPEs of 14.69\% and 12.58\% for energy and latency, respectively. The communication and memory transpose kernel patterns are validated with Torch profiler.

\input{elements/llm_spec_cp}

\clearpage
\newpage
\section{Additional Plots for Llama3-70B with TP}
\label{sec:app_llama_decode}

\begin{figure}[!h]
    \centering
    \begin{subfigure}[b]{0.48\linewidth}
    \centering
    \includegraphics[width=\linewidth]{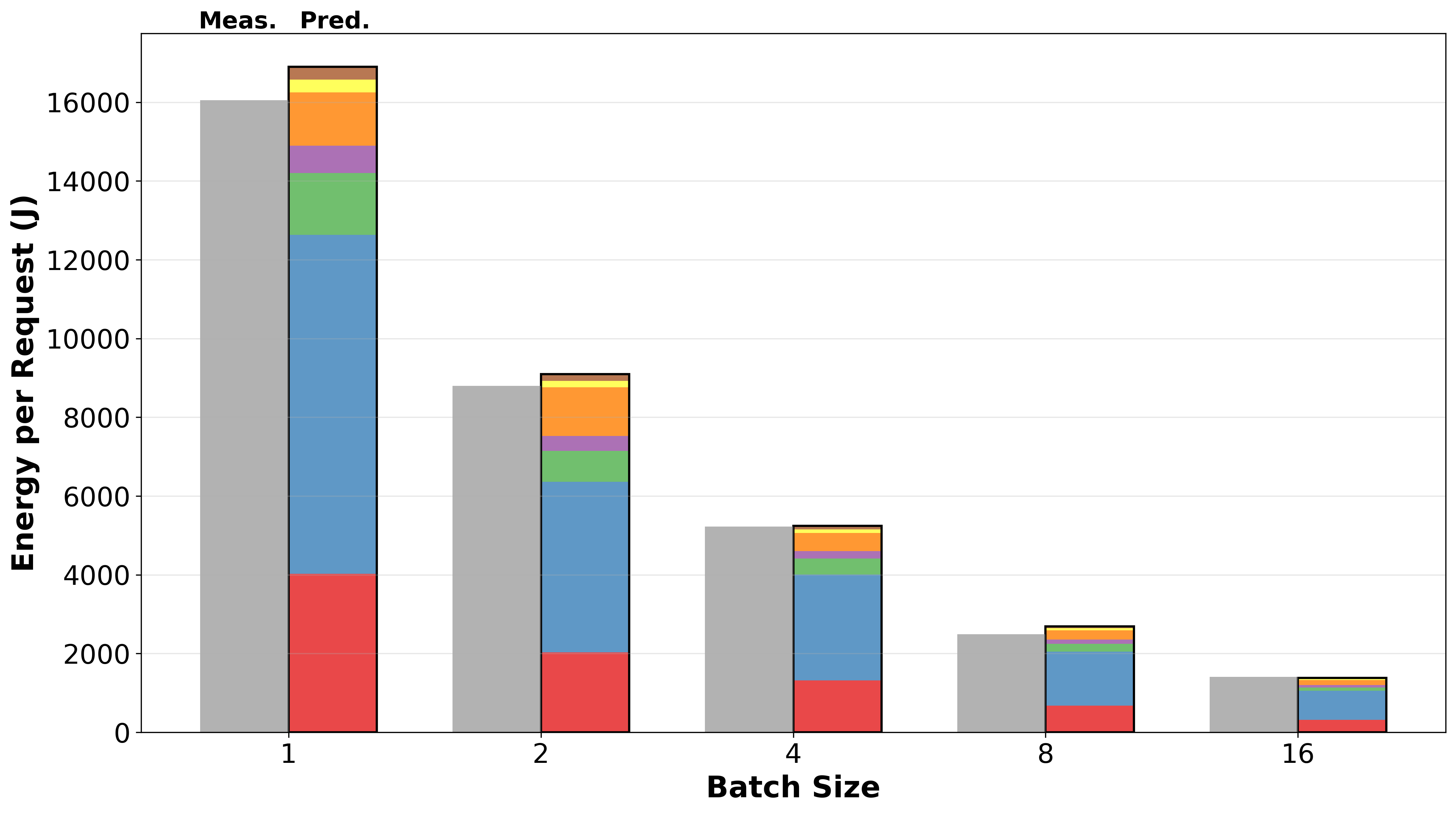}
    \caption{Decode energy breakdown with the same legend as \Cref{fig:70B_prefill_E}.}

    \label{fig:70B_decode_E}

    \end{subfigure}
    \begin{subfigure}[b]{0.48\linewidth}
    \centering
    \includegraphics[width=\linewidth]{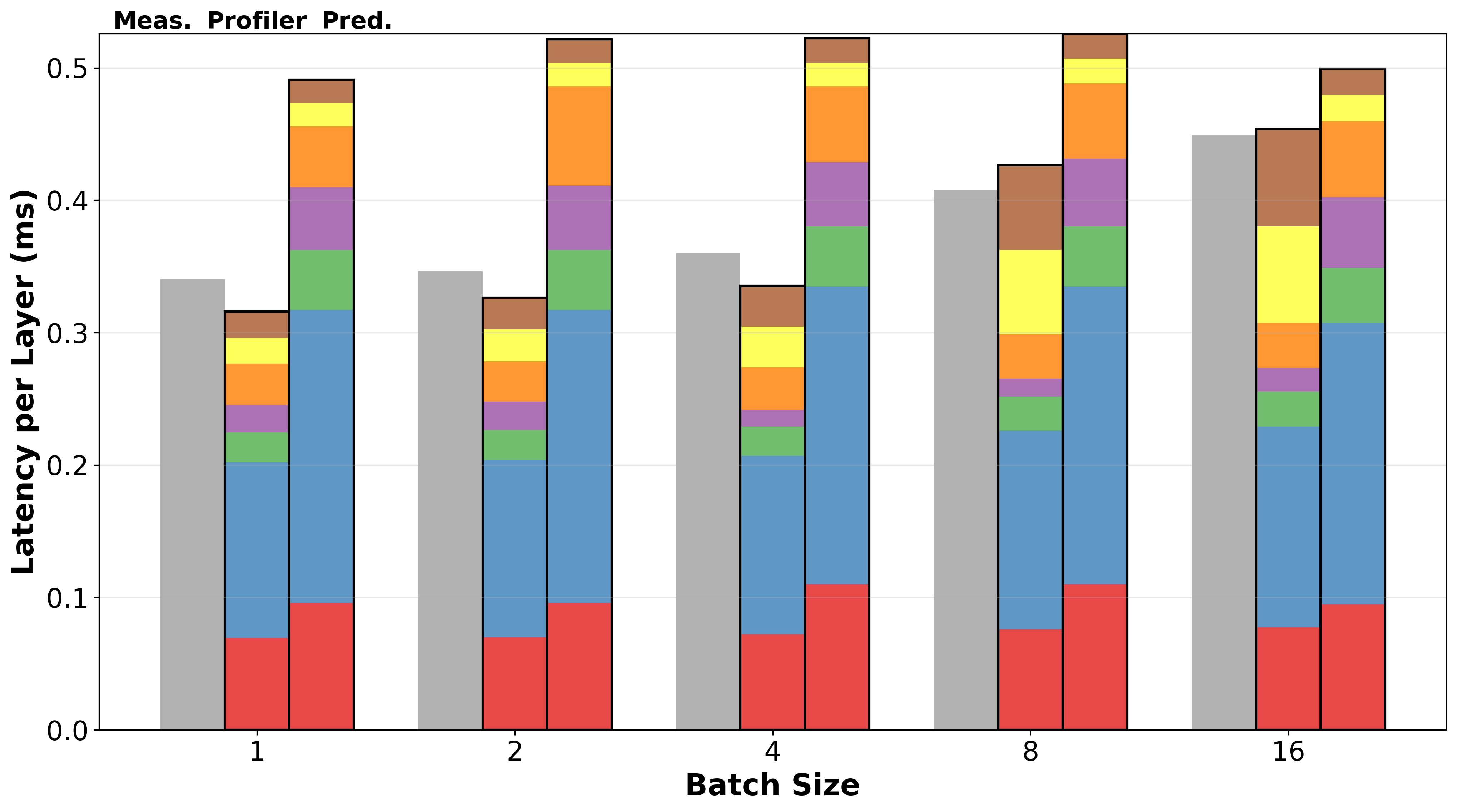}
    \caption{Decode latency breakdown with the same legend as \Cref{fig:70B_prefill_E}.}

    \label{fig:70B_decode_T}
\end{subfigure}

 \begin{subfigure}[b]{0.48\linewidth}
    \input{elements/kv_tp4}
    \caption{Memory constraint.}
    \label{fig:70B_decode_memory}
\end{subfigure}
\hfill
\begin{subfigure}[b]{0.3\linewidth}
    \includegraphics[width=\linewidth]{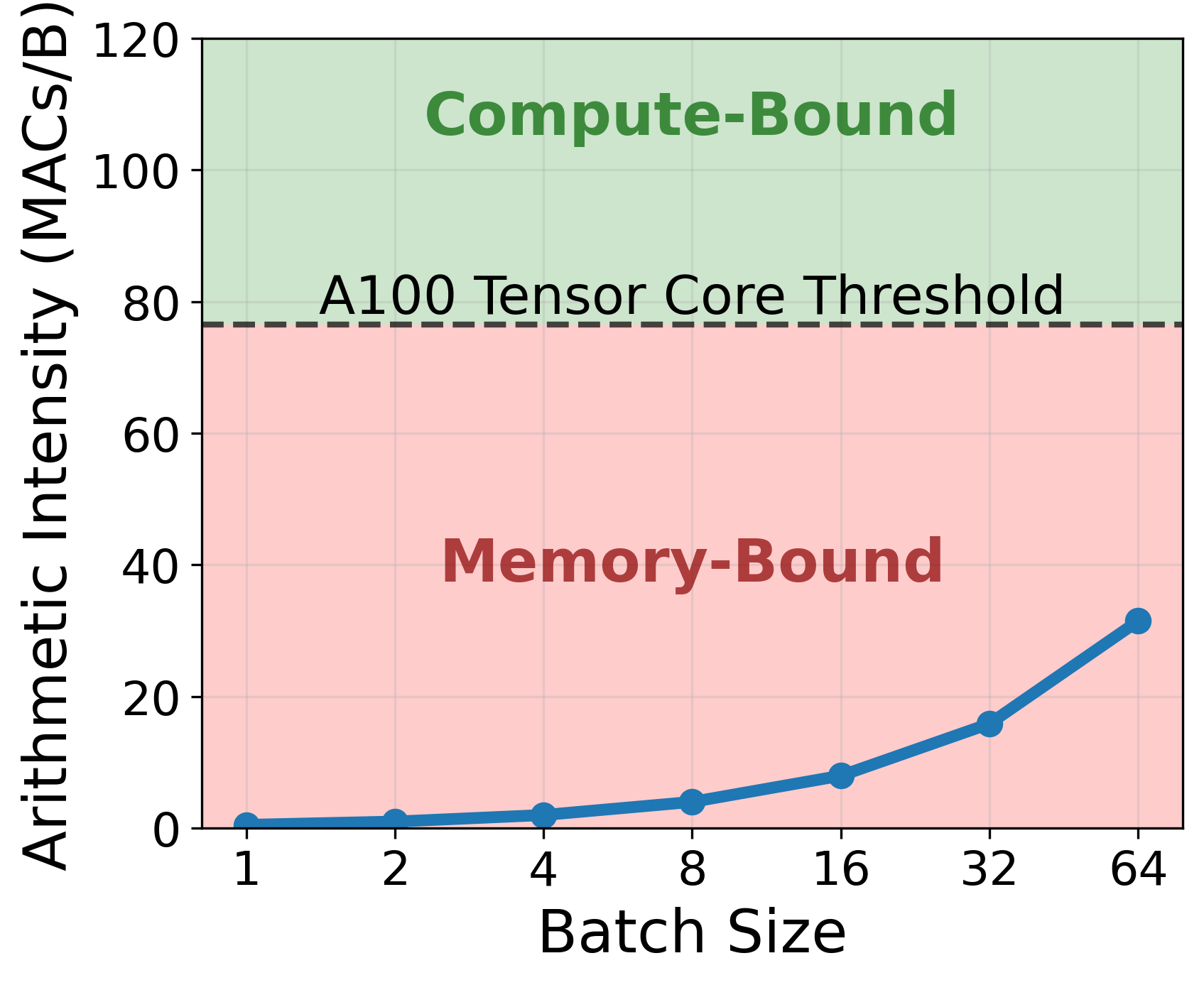}
    \caption{Down projection.}
    \label{fig:CI}
\end{subfigure}

{\small
In all breakdown figures, the profiled latency breakdown is from the Torch traces of the first 10 decoding steps, instead of the full \ac{OSL}=1024. 
}
    \caption{Energy and latency breakdown for Llama3-70B decode phase at TP4. (a) Energy breakdown shows the benefit of batching. (b) Latency breakdown shows limitations of kernel-level latency prediction for small decode operations.}
    \label{fig:70B_decode_combined}

\end{figure}

In the decode phase, energy consumption varies more significantly with batch size than with tensor parallelism (\Cref{fig:dse_decode}). However, energy savings trade off with memory requirements. Increase in batch size means only shorter sequences can fit inside the GPU DRAM (\Cref{fig:70B_decode_memory}). During decode, operations exhibit skewed tensor shapes that result in low arithmetic intensity.
For example, \ourwork{} parses the down projection \texttt{op("bsf,fm->bsm", parallel="f")} into a standard matmul operation with output dimensions $M=b$ and $N=m$ (since $s=1$ in decode).
With $M$ equivalent to the batch size, the GPU's \acp{SM}, the compute units, remain underutilized due to being memory-bound (\Cref{fig:CI}). Scaling up batch size improves arithmetic intensity, making it more efficient.

Across parallelism configurations (TP2, TP4, and TP8), batch sizes (1 to 16), and \ac{OSL} up to 7680 tokens, \ourwork{} achieves a \ac{MAPE} of \decodemultiE{} for decode energy. Decode latency exhibits larger errors (MAPE=\decodemultiT{}). The latency breakdown in \Cref{fig:70B_decode_T} reveals limitations of kernel-level latency prediction for low-arithmetic-intensity operations and skewed matrix shapes inherent to small decode operations. We verified this limitation persists across alternative compute kernel backends \citep{neusight,li2023}, confirming it is not backend-specific.
Despite higher latency errors, \ourwork{} correctly captures that decode latency remains relatively independent of batch size.

\clearpage
\newpage
\section{Additional Plots for Qwen3-30B-A3B with TP+EP}
\label{sec:app_qwen_prefill}
\begin{figure}[!htbp]
    \centering
        \begin{subfigure}[b]{0.48\linewidth}
        \centering
    \includegraphics[width=\linewidth]{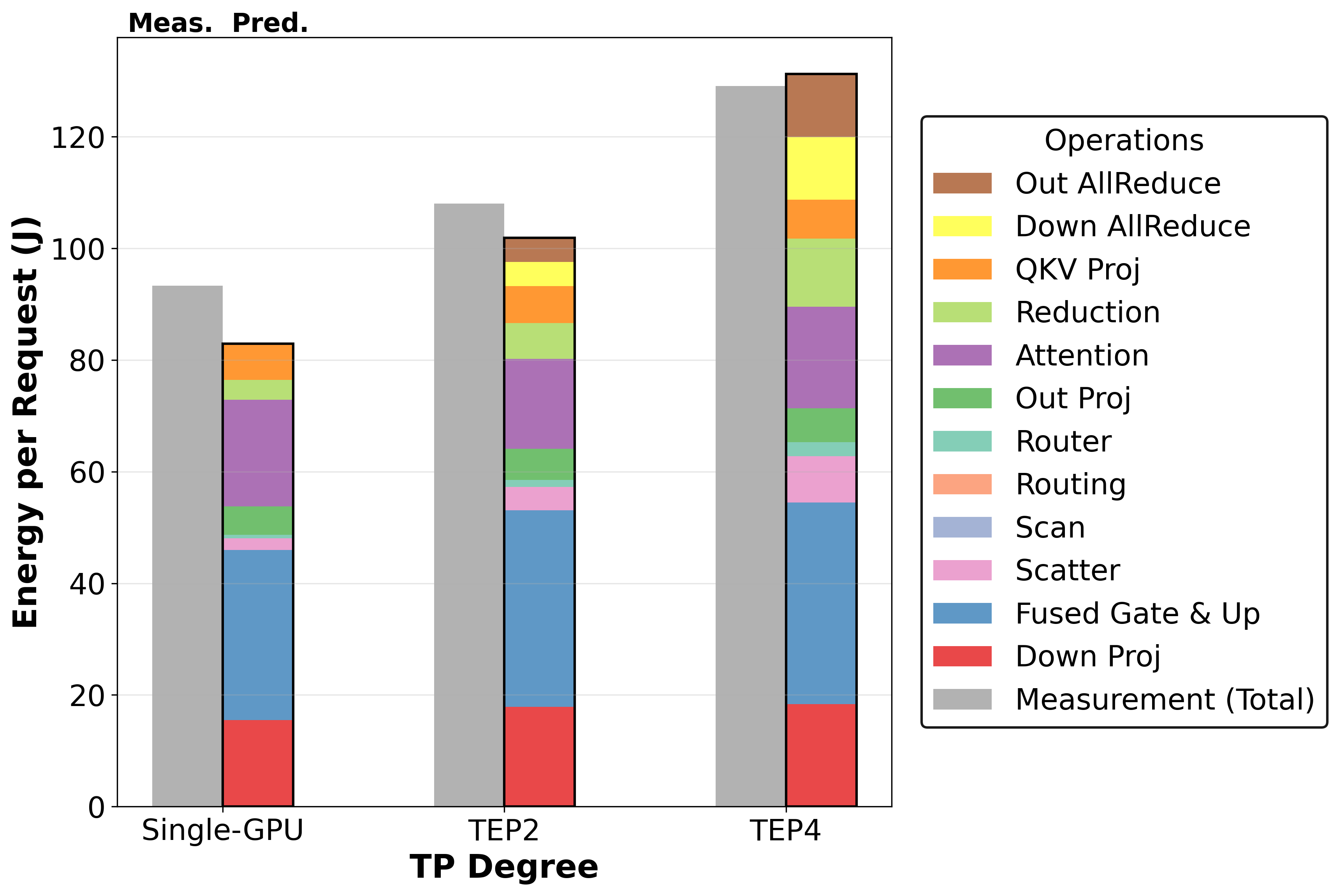}
    \caption{Prefill energy breakdown.}
        \label{fig:qwen_prefill_E}
    \end{subfigure}
    \begin{subfigure}[b]{0.48\linewidth}
        \centering
    \includegraphics[width=\linewidth]{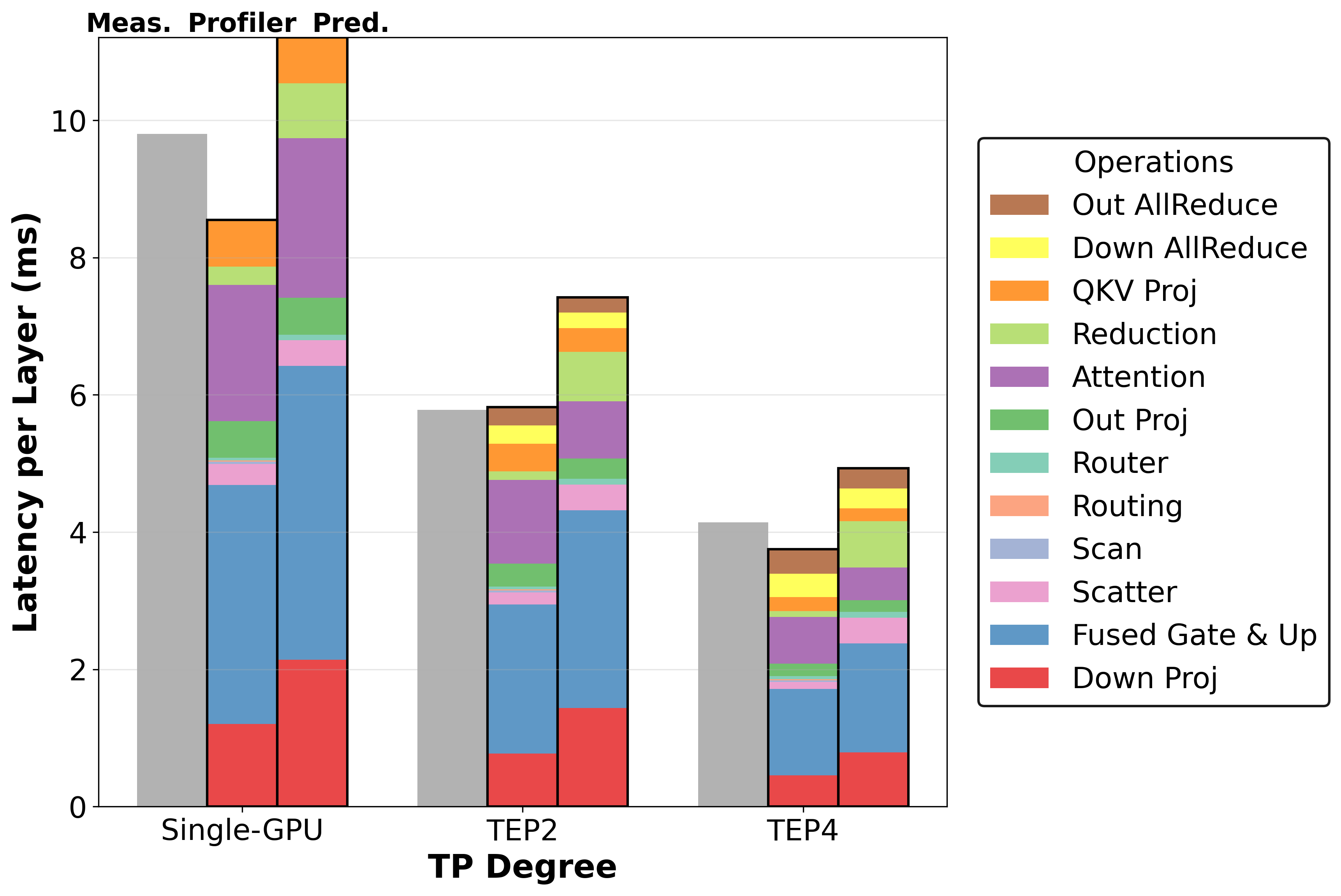}
    \caption{Prefill latency breakdown.}
    \label{fig:qwen_prefill_T}
    \end{subfigure}
    \caption{Qwen3-30B-A3B prefill phase at B2, ISL=4096 across \ac{TP} in attention and \ac{EP} in MoE, demonstrating increasing communication overhead as TP degree increases.}
    \label{fig:qwen_prefill}
\end{figure}
Qwen3-30B-A3B is evaluated with TP + EP up to 4 GPUs, because there are only 4 key-value heads in the attention layer.
The measured results are carefully calibrated to contain warmup iterations and measured at the steady state. However, the profiled breakdown results are averaged across all the layers of 1 iteration for prefill and 10 iterations for decode. Similarly to the dense model, we observe that increased TP degree leads to increased communication cost.

\clearpage
\newpage
\section{Evaluation of Kernel Fusion}
\label{sec:app_fusion}
\begin{figure}[h]
    \centering
    \includegraphics[width=0.8\linewidth]{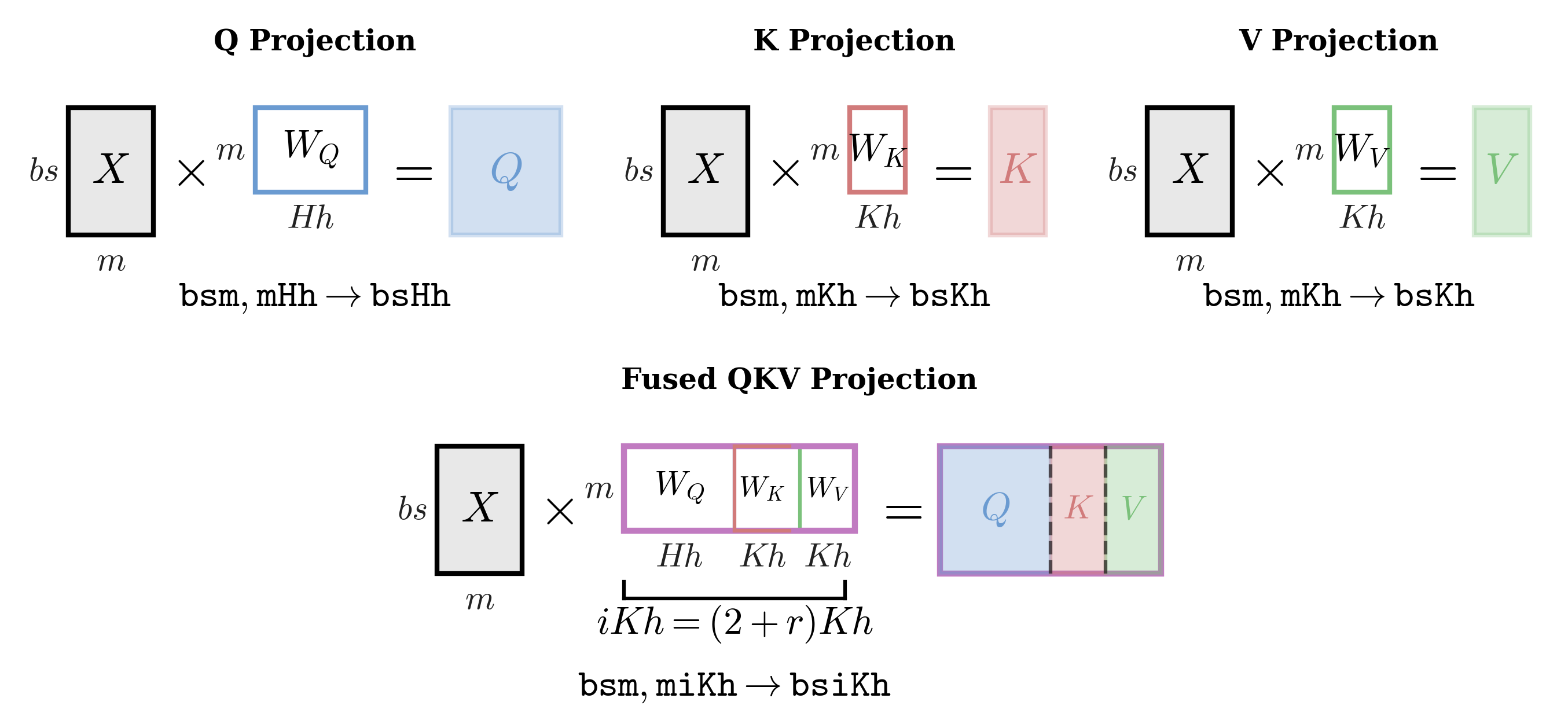}
    \caption{\ourwork{} allows developers to rapidly assess unfused and fused implementations. Example shown here for the query (Q), key (K), and value (V) projections.}
    \label{fig:fusion}
\end{figure}

A key advantage of EnergyLens is that it enables rapid exploration of fusion strategies across the full range of serving scenarios without requiring implementation or profiling. Instead, the developers can specify any range of \ac{ISL}, \ac{OSL}, batch sizes, and tensor parallelism configurations for evaluation. The fused and unfused Llama3 specifications are in \Cref{lst:ops} and \Cref{lst:ops_unfused}, with the fused QKV projections shown in \Cref{fig:fusion}.

\begin{figure}[htbp]
    \centering
    \begin{subfigure}[b]{0.8\linewidth}
        \centering
        \includegraphics[width=\linewidth]{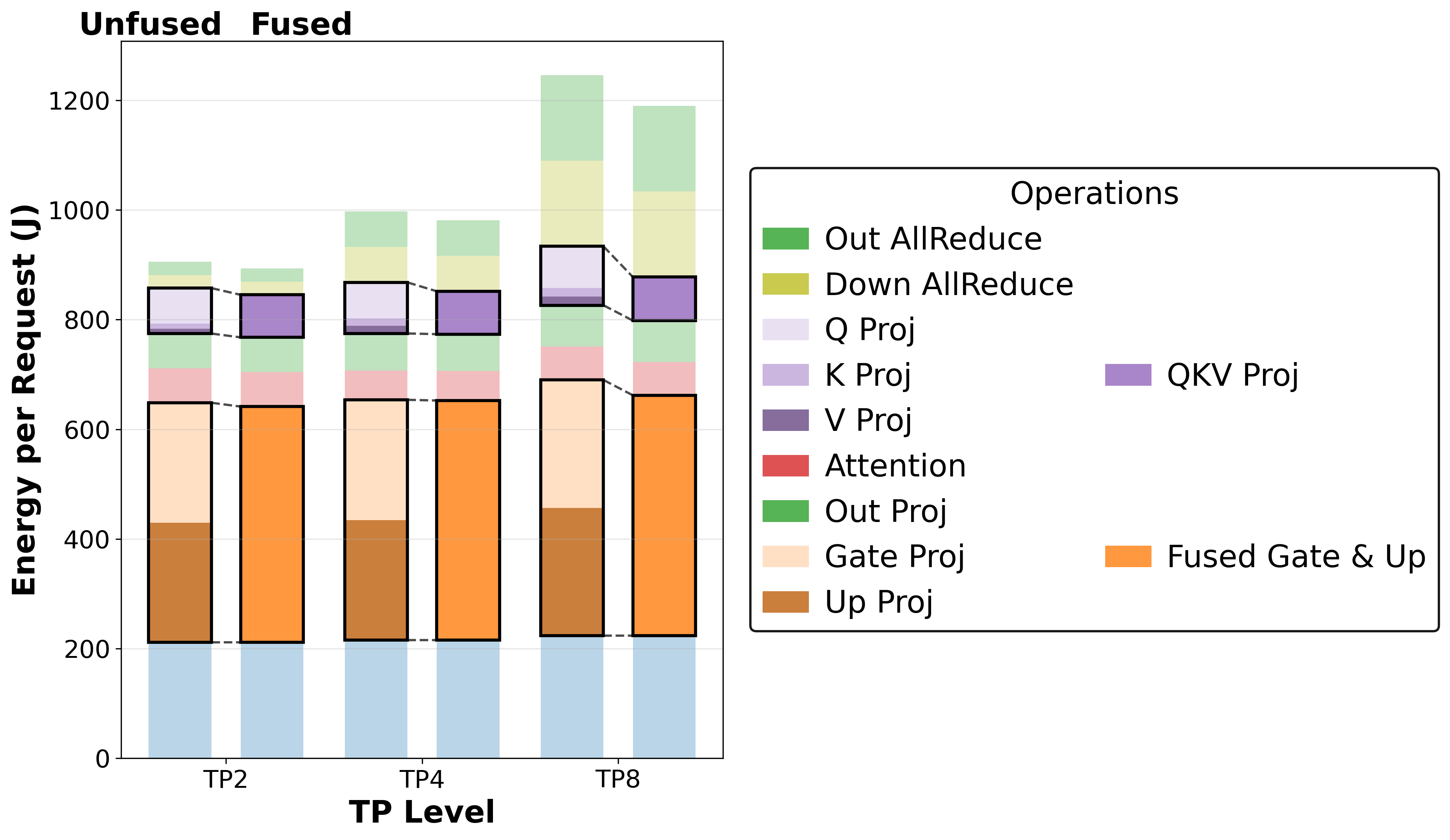}
        \caption{Prefill phase (B2, ISL=4096).}
        \label{fig:fusion_prefill}
    \end{subfigure}
    \vspace{0.5em}
    
    \begin{subfigure}[b]{0.8\linewidth}
        \centering
        \includegraphics[width=\linewidth]{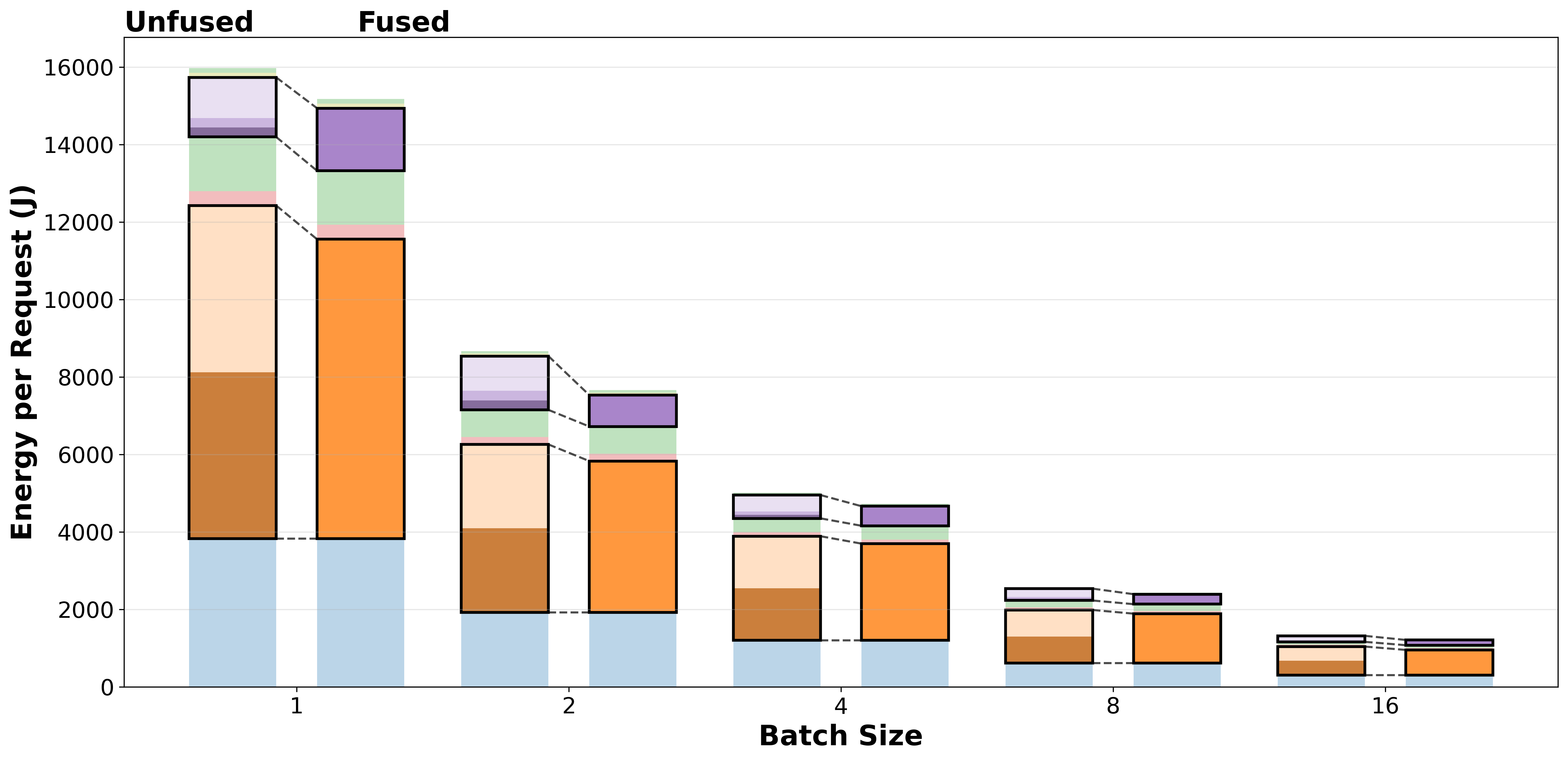}
        \caption{Decode phase (TP2, ISL=512, OSL=1024), same legend as (a).}
        \label{fig:fusion_decode}
    \end{subfigure}
    \caption{\ourwork{} enables fast evaluation of kernel fusion.}
    \label{fig:fusion_comparison}
\end{figure}

We demonstrate this capability by evaluating the energy impact of layer fusion across example deployment settings in \Cref{fig:fusion_comparison}. The percentage energy saving in the prefill phase, averaged across batch sizes (1 to 64), ISL (256 to 8192), and TP (2,4,8) is \fuseprefill{}. The energy breakdown reveals that fusing Q, K, V projections yields substantial savings by reducing memory traffic. This benefit is particularly pronounced in the decode phase where operations exhibit lower arithmetic intensity and are more sensitive to kernel overhead. At $ISL=512$, we observe an average energy saving of \fusedecode{} for decode phase workloads.

\clearpage
\newpage
\section{Limitations}
\label{sec:limit}
\input{elements/backend}
In the decode phase of LLM inference, \ac{GEMM} kernels exhibit low arithmetic intensity and skewed matrix shapes that challenge existing kernel latency estimation tools \citep{li2023,neusight,energaizer}. To assess whether this limitation stems from our default backend, we leverage EnergyLens's modular design to evaluate alternative backends. As shown in Table~\ref{tab:backend_comparison}, NeuSight yields comparable decode latency errors, confirming that the challenge is inherent to small, low-intensity operations rather than backend-specific. Custom kernel libraries (e.g., TensorRT-LLM) may achieve superior performance compared to cuBLAS-dispatched kernels, contributing to discrepancies observed in \Cref{fig:70B_decode_T}. For communication kernels, we observe high consistency with measurements at large transfer sizes, though small discrepancies appear at smaller sizes.

\section{Accuracy}
\ourwork{} focuses on evaluating the energy efficiency of \ac{LLM} inference, achieving energy \ac{MAPE}s of \prefillmultiE{} and \decodemultiE{} for prefill and decode phases respectively. These prediction errors are substantially smaller than the energy variations across different deployment configurations observed in \Cref{sec:dse}, validating \ourwork{}'s utility for energy-driven optimization.

Beyond energy prediction, \ourwork{} supports latency estimation to enable energy-latency trade-off analysis. While achieving high accuracy during prefill (MAPE=\prefillmultiT{}), latency prediction exhibits larger errors during decode (MAPE=\decodemultiT{}). This limitation causes configurations B4 and B8 in \Cref{fig:dse_decode} to appear slightly slower than measured. Despite this discrepancy, \ourwork{} correctly captures the behavior that decode latency remains relatively independent of batch size.

\clearpage
\newpage
\section{Reproducibility}
\label{sec:app_reproduce}
We reran the Llama3-70B \ac{TP} experiments. This occurred because our initial experiment data was accidentally wiped. We report the old and new \acp{MAPE} below. All \acp{MAPE} closely align with the original, showing that our experimental setup, including the profiling process, is reproducible.
\input{elements/reproducible}

\clearpage
\newpage
\section{Experimental Setup and Compute Resources}
\label{sec:app_setup}
Here we provide details in addition to \Cref{sec:eval_method}. All models are evaluated in bf16 precision. For Llama3 models, we use TensorRT-LLM v0.14. \Cref{lst:ops} represents it in the \ourwork{} interface. Qwen3-30B-A3B experiments were done with TensorRT-LLM v1.0 and \Cref{lst:qwen} describes the fusion and parallel strategy. This combination of \ac{TP} and \ac{EP} is termed as \texttt{tep} in TensorRT-LLM documentation. The workload used in profiling is from Hugging Face and can be obtained with \texttt{load\_dataset("wikipedia", "20220301.en", split="train")}.

For all prefill MAPE results, the evaluation swept input sequence length linearly spaced with interval of 256 up to 8192 and batch size exponentially spaced up to 64 with a factor of 2. For all decode MAPE results, the evaluation used ISL 512, swept output sequence length linear spaced up to OSL 7680 or until out-of-memory if earlier, and swept batch size up to 16. Actual runtime batch sizes used by TensorRT-LLM at long contexts were verified with Torch Profiler, and all observed batch-size/sequence-length pairs were included in the MAPE calculation.

The Llama3-70B overlap MAPE results were obtained with Megatron-style compute-communication overlap in the prefill phase. Overlap configurations including no overlap and 4-stage overlap with 1, 4, or 16 SMs dedicated to communication.

All Llama3 TP experiments include TP2, TP4, TP8. All Qwen experiments include TP+EP2 and TP+EP4 (limited by the number of key-value heads).

The main cost of the experiments comes from validating the accuracy by actually running the workloads. We provide a table of the GPU-hour cost below. We also estimate 200 GPU-hour of auxiliary experiments, mainly collecting execution traces to validate our model.

\input{elements/gpuhour}

\subsection{Licenses}
\label{sec:app_license}
We use the following existing assets in our experiments: TensorRT-LLM v0.14 and v1.0 (https://github.com/NVIDIA/TensorRT-LLM, license: Apache 2.0); Megatron-LM v0.15.3 (https://github.com/NVIDIA/Megatron-LM, license: BSD 3-Clause); PyTorch / Torch Profiler (https://github.com/pytorch/pytorch, license: BSD-style); NCCL microbenchmarks (https://github.com/NVIDIA/nccl-tests, license: BSD 3-Clause "New" or "Revised" License); Llama 3 models (https://github.com/meta-llama/llama3, license/terms: Meta Llama 3 Community License Agreement); Qwen3-30B-A3B (https://huggingface.co/Qwen/Qwen3-30B-A3B, license/terms: Apache 2.0); EnergAIzer (https://zenodo.org/records/18916559, MIT license). We cite the corresponding papers and repositories and use these assets in accordance with their stated terms.

%% file: elements/llm_spec_unfused.tex
\begin{listing}[th]
\begin{minted}[
    fontsize=\small,
    linenos,
    breaklines,
    breaksymbolright=,
    xleftmargin=1em,
    breaksymbolleft=\hspace{2em},
    escapeinside=||
]{python}
llm_model.EQS = [
    op("bsm,mHh->bsHh", parallel="H", label="Q Projection"),|\label{line:unfus_q}|
    op("bsm,mKh->bsKh", parallel="K", label="K Projection"),|\label{line:unfus_k}|
    op("bsm,mKh->bsKh", parallel="K", label="V Projection"),|\label{line:unfus_v}|
    op("attention", parallel="H", label="Attention"),
    op("bsHh,Hhm->bsm", parallel="H", label="Output Projection"),
    op("bsm,mf->bsf", parallel="f", label="Gate Projection"),|\label{line:unfus_gate}|
    op("bsm,mf->bsf", parallel="f", label="Up Projection"),|\label{line:unfus_up}|
    op("bsf,fm->bsm", parallel="f", label="Down Projection")
]
\end{minted}
\caption{Full specification of an unfused Llama3 implementation.}
\label{lst:ops_unfused}
\end{listing}

%% file: elements/llm_spec_fused.tex
\begin{listing}[th]
\begin{minted}[
    fontsize=\small,
  %  frame=lines,
    linenos,                    % Enable line numbers
    xleftmargin=1em,
    breaklines,
    breaksymbolright=,
    breaksymbolleft=\hspace{2em},
    escapeinside=||             % Allows LaTeX commands inside ||
]{python}
llm_model.EQS = [
    op("bsm,miKh->bsiKh", parallel="K", label="QKV Projection"),|\label{line:fus1}|
    op("attention", parallel="K", attn_eqs=attn_eqs, label="Attention"),
    op("bsHh,Hhm->bsm", parallel="H", label="Output Projection"),
    op("bsm,mF->bsF", parallel="F", label="Gate & Up Projection"),|\label{line:fus2}|
    op("bsf,fm->bsm", parallel="f", label="Down Projection")|\label{line:down}|
]
\end{minted}
\caption{Full specification of a fused Llama3 implementation.}
\label{lst:ops}
\end{listing}

%% file: elements/attn_eq.tex
\begin{listing}[th]
\begin{minted}[
    fontsize=\small,
    linenos,
    xleftmargin=1em,
    breaklines,
    breaksymbolright=,
    breaksymbolleft=\hspace{2em},
    escapeinside=||
]{python}
attn_eqs = [
    op("bKrsh,bKzh->bKrsz", parallel="K", label="QK"),|\label{line:qk}|
    op("bKrsz,bKzh->bKrsh", parallel="K", label="AV")|\label{line:av}|
]
\end{minted}
\caption{Grouped query attention sub-equations with attention-head-parallel used in \Cref{lst:qwen,lst:ops,lst:ops_unfused}.}
\label{lst:attn_eqs}
\end{listing}

%% file: elements/llm_sym.tex
\begin{table}[ht]
\centering
\caption{LLM symbols used in \Cref{lst:ops,lst:qwen,lst:ops_unfused}}
\label{tab:notation}
\begin{tabular}{cl}
\hline
Symbol & Description \\
\hline
$r$ & Q head to KV head ratio in grouped query attention \\
$b$ & Batch size \\
$s$ & Input length in prefill; 1 in decode \\
$h$ & Head dimension \\
$H$ & Number of query heads \\
$K$ & Number of key/value heads \\
$t$ & Bytes per data type \\
$m$ & $d_{model}$, i.e. embeddings or hidden dimension \\
$f$ & FFN dimension \\
$z$ & Input length in prefill; input + output length in decode \\
$i$ & $2+r$, used in the fused QKV projection \\
$A$ & Number of activated experts per token \\
$F$ & $2f$, used in the fused gate \& up projection \\
\hline
\end{tabular}
\end{table}

%% file: elements/allreduce.tex
\begin{algorithm}
\caption{AllReduce Collective Kernel Detection}
\label{alg:allreduce}
\begin{algorithmic}[1]
\Function{NeedAllReduce}{op}
    \State $\textit{summation\_indices} \gets \Call{ParseEinsum1}{\text{op.eq}}$
    \If{$\text{op.parallel} \in \textit{summation\_indices}$}
        \State \Return \textbf{true}
    \Else
        \State \Return \textbf{false}
    \EndIf
\EndFunction

\Statex

\If{\Call{NeedAllReduce}{$\textit{op}$}}
\State $\textit{output} \gets \Call{ParseEinsum2}    {\text{op.eq}}$
    \State $\textit{size} \gets \Call{GetMatrixSize}{\textit{output}, \text{self.params}} \times t $
    \State $\textit{all\_reduce}.\Call{Append}{\textit{size}}$
\Else
    \State $\textit{all\_reduce}.\Call{Append}{0}$
\EndIf
\end{algorithmic}

\end{algorithm}

%% file: elements/all2all.tex
\begin{algorithm}
\caption{All-to-All Collective Kernel Detection}
\label{alg:all2all}
\begin{algorithmic}[1]
\Function{NeedAll2All}{op, prev\_op}
    \If{$\text{op.CP\_dim} \neq \text{prev\_op.CP\_dim}$}
        \State \Return \textbf{true}
    \Else
        \State \Return \textbf{false}
    \EndIf
\EndFunction
\Statex
\If{\Call{NeedAll2All}{$\textit{op}, \textit{prev\_op}$}}
    \State $\textit{output} \gets \Call{ParseEinsum2}{\text{prev\_op.eq}}$
    \State $\textit{size} \gets \Call{GetMatrixSizePerRank}{\textit{output}, \text{self.params}}$
    \State $\textit{kernels}.\Call{Append}{\textit{transpose}, \textit{size}}$ \Comment{Continuous tensor layout required for NCCL}
    \State $\textit{kernels}.\Call{Append}{\textit{all2all}, \textit{size}}$
    \State $\textit{kernels}.\Call{Append}{\textit{transpose}, \textit{size}}$
\EndIf
\end{algorithmic}
\end{algorithm}

%% file: elements/llm_spec_cp.tex
\begin{listing}[th]
\begin{minted}[
    fontsize=\small,
    linenos,
    xleftmargin=1em,
    breaklines,
    breaksymbolright=,
    breaksymbolleft=\hspace{2em},
    escapeinside=||
]{python}
attn_eqs = [
    op("bKrsh,bKzh->bKrsz", CP_dim="K", label="QK"),|\label{line:qk}|
    op("bKrsz,bKzh->bKrsh", CP_dim="K", label="AV")|\label{line:av}|
]
llm_model.EQS = [
    op("bsm,mi->bsi", CP_dim="s", label="QKV Projection"),|\label{line:fus1}|
    op("attention", CP_dim="H", label="Attention"),
    op("bsHh,Hhm->bsm", CP_dim="s", label="Output Projection"),
    op("bsm,mF->bsF", CP_dim="s", label="Gate & Up Projection"),|\label{line:fus2}|
    op("bsf,fm->bsm", CP_dim="s", label="Down Projection")|\label{line:down}|
]
\end{minted}
\caption{Full specification of a fused Llama3 implementation with context parallelism.}
\label{lst:ops_cp}
\end{listing}

%% file: elements/kv_tp4.tex
\begin{center}
{\small
\begin{tabular}{cc}
\toprule
\textbf{Batch} & \textbf{Max Seq Length} \\
\midrule
1 & 147,456 \\
2 & 73,728 \\
4 & 36,864 \\
8 & 18,432 \\
16 & 9,216 \\
32 & 4,608 \\
64 & 2,304 \\
\bottomrule
\end{tabular}
}
\end{center}

%% file: elements/backend.tex
\begin{table}[h]
\centering
\small
\caption{Decode latency MAPE across different kernel-level modeling backends for Llama3-70B.}
\label{tab:backend_comparison}
\begin{tabular}{lc}
\toprule
Latency Backend Tool & Decode Latency MAPE \\
\midrule
EnergAIzer \citep{energaizer} & \decodemultiT \\
\cite{li2023} & \decodemultiTLi \\
NeuSight \citep{neusight} & \decodemultiTNeusight \\
\bottomrule
\end{tabular}
\end{table}

%% file: elements/reproducible.tex
\begin{table}[h]
\centering
\caption{Comparison of original and reproduced MAPE values for Llama3-70B \ac{TP} experiments.}
\label{tab:reproducibility}
\begin{tabular}{llcc}
\toprule
\textbf{Phase} & \textbf{Metric} & \textbf{Original} & \textbf{Reproduced} \\
\midrule
Prefill  & Energy  & \oldprefillmultiE & \prefillmultiE \\
Prefill  & Latency & \oldprefillmultiT & \prefillmultiT \\
Decode   & Energy  & \olddecodemultiE  & \decodemultiE  \\
Decode   & Latency & \olddecodemultiT  & \decodemultiT  \\
\bottomrule
\end{tabular}
\end{table}

%% file: elements/gpuhour.tex
\begin{table}[ht]
  \centering
  \caption{Aggregate GPU-hours by experiment group.}
  \label{tab:gpu-hours}
  \small
  \begin{tabular}{@{}lr@{}}
    \toprule
    \textbf{Experiment Group} & \textbf{GPU-hours} \\
    \midrule
    Llama3-70B TensorRT-LLM        & 265  \\
    Llama3-70B Megatron (overlap)   & 389  \\
    Llama3-8B                       & 21   \\
    Qwen-30B-A3B                    & 119  \\
    NCCL microbenchmarks            & 60   \\
    Auxiliary (trace collection)    & 200  \\
    \midrule
    \textbf{Total}                  & \textbf{1\,054} \\
    \bottomrule
  \end{tabular}
\end{table}

%% file: elements/abr.tex
\begin{acronym}[GEMM]
\acro{TTFT}{time to first token}
\acro{ETFT}{energy to first token}
\acro{LLM}{large language model}
\acro{TP}{tensor parallelism}
\acro{MAPE}{mean absolute percentage error}
\acro{EPOT}{energy per output token}
\acro{TPOT}{time per output token}
\acro{ISL}{input sequence length}
\acro{OSL}{output sequence length}
\acro{FFN}{feed-forward network}
\acro{GEMM}{general matrix multiplication}
\acro{SM}{streaming multiprocessor}
\acro{TDP}{thermal design power}
\acro{NVML}{NVIDIA Management Library}
\acro{EP}{expert parallelism}
\acro{CP}{context parallelism}
\acro{MoE}{mixture of experts}
\end{acronym}